%% file: acl_latex.tex
\pdfoutput=1

\documentclass[11pt]{article}

\usepackage{acl}

\usepackage{times}
\usepackage{latexsym}

\usepackage[T1]{fontenc}

\usepackage[utf8]{inputenc}

\usepackage{microtype}

\title{R-Spin: Efficient Speaker and Noise-invariant\\Representation Learning with Acoustic Pieces}

\author{Heng-Jui Chang \and James Glass \\
        MIT CSAIL\\
        \texttt{hengjui@mit.edu}\\}

\usepackage{multirow}
\usepackage{threeparttable}
\usepackage{subcaption}
\usepackage{booktabs}
\usepackage{graphicx}
\usepackage{amssymb,amsmath}
\usepackage{adjustbox}
\usepackage{enumitem}
\usepackage{bbding}
\usepackage{stfloats}

\newcommand{\proposed}{R-Spin}
\newcommand{\transp}{\mathsf{\scriptscriptstyle{T}}}

\begin{document}
\maketitle
\begin{abstract}
This paper introduces Robust Spin~(R-Spin), a data-efficient domain-specific self-supervision method for speaker and noise-invariant speech representations by learning discrete acoustic units with speaker-invariant clustering~(Spin).
R-Spin resolves Spin's issues and enhances content representations by learning to predict acoustic pieces.
R-Spin offers a 12X reduction in computational resources compared to previous state-of-the-art methods while outperforming them in severely distorted speech scenarios.
This paper provides detailed analyses to show how discrete units contribute to speech encoder training and improving robustness in diverse acoustic environments.
\end{abstract}

\input{section/intro}
\input{section/method}
\input{section/exp}
\input{section/conclusion}
\input{section/ack}
\input{section/limitation}
\input{section/ethics}

\bibliography{anthology,custom}
\bibliographystyle{acl_natbib}

\clearpage

\input{section/appendix}

\end{document}

%% file: section/intro.tex
\section{Introduction}
\label{sec:intro}

Self-supervised learning~(SSL) for encoder pre-training has emerged as a foundational element in speech processing, outperforming conventional approaches across various applications~\citep{mohamed2022self,liu2022audio}.
Given the substantial cost associated with human annotation of speech data, SSL methods leverage unlabeled audio data to pre-train encoders, generating good representations for downstream tasks like automatic speech recognition~(ASR) and speaker identification~\citep{yang2021superb,tsai-etal-2022-superb}.
The application of SSL models has notably concentrated on ASR, aiming to mitigate the dependence on large transcribed corpora~\citep{hsu2021hubert,baevski2022data2vec,liu2023dinosr}.
Thus, extracting content representations has become a crucial aspect of speech SSL research~\citep{tjandra2020unsupervised,chan2022content,peyser2022towards,williams2022learning}.
Prior studies have devised objective functions to disentangle content from speech, fostering the ability of SSL models to generate speaker-invariant representations through domain-specific self-supervision~(DS).
In DS, a pre-trained SSL model is fine-tuned with unlabeled data for specific applications.
\citet{qian2022contentvec} propose ContentVec by disentangling speaker and content information, demonstrating promising results.
However, ContentVec suffers from the requirement of a voice conversion model and substantial computational costs exceeding 600 GPU hours.
Alternatively, \citet{chang2023spin} propose Speaker-invariant Clustering~(Spin) to produce content representations with minimal fine-tuning resources.
Nonetheless, Spin is constrained to fine-tuning only the top layers, thereby lacking the flexibility to adapt to diverse acoustic domains.

\input{fig/framework}

Parallel to modeling content information in speech, numerous studies are dedicated to investigating the robustness of speech SSL.
While current methods perform well on clean speech datasets, they are vulnerable to out-of-domain data like distorted audio signals~\citep{hsu2021robust}.
To mitigate this vulnerability, researchers have proposed noise-invariant training techniques.
\citet{huang2022improving} proposes HuBERT-MGR via domain adversarial training to render the fine-tuned HuBERT model~\citep{hsu2021hubert} invariant to domain shifts.
WavLM~\citep{chen2022wavlm} integrates denoising with the HuBERT pre-training framework, achieving state-of-the-art performance in many speech processing downstream tasks.
Similarly, \citet{zhu2023robust} propose Robust data2vec, introducing perturbations to the input to predict the exponential moving average teacher model's representations.
In deHuBERT~\citep{ng2023dehubert}, the Barlow Twins loss~\citep{zbontar2021barlow} is applied to encourage representation invariability to input perturbations.
Although many methods have shown success in noisy speech recognition~\citep{wang2022wav2vec-switch,zhu2022noise,huang2022spiral,hu2023wav2code}, to our knowledge, none have concurrently addressed the disentanglement of speaker and noise while enhancing content information.
Furthermore, these approaches exhibit inefficiency, often requiring high computation costs and iterating large corpora over numerous epochs.

To effectively acquire high-quality content and robust representations for real-world applications, this paper extends Spin with noise-invariant training and acoustic piece pseudo-label learning, coined Robust Spin~(\proposed).
During training, two utterances of the same content with different distortions are fed into a speech SSL encoder.
The outputs are frame-wise vector-quantized with a learnable codebook via online clustering, as in Spin.
The model is trained to match cluster ID distributions between the utterances.
To prevent codebook collapse, an additional pseudo-label prediction loss is introduced.
The pseudo-labels are generated by learning acoustic pieces~\citep{ren2022acoustic-piece} on top of the discrete units produced by a pre-trained Spin model, offering better training targets that closely align with phonemes and characters.
Within this framework, the speech encoder learns speaker and noise-invariant representations, benefiting robustness and content extraction simultaneously.
The contributions are summarized as follows:
\begin{enumerate}[nolistsep]
    \item We integrate predicting acoustic pieces into Spin, enabling fine-tuning all parameters without collapsing, which allows the processing of more complex speech recordings.
    \item \proposed~inherits the benefit of low training costs from Spin, requiring 12X less computation than prior art.
    \item With noise-invariant training, \proposed~outperforms other DS approaches in distorted speech and phoneme recognition tasks like the CHiME-4 challenge~\citep{vincent2017chime4}.
    \item We inspect the hidden representations of speech SSL models to quantify the speaker and noise invariability.
    \item We offer in-depth analyses of discrete acoustic units to understand how these units help speech encoder training.
\end{enumerate}

%% file: fig/framework.tex
\begin{figure*}
    \centering
    \includegraphics[width=0.9\linewidth]{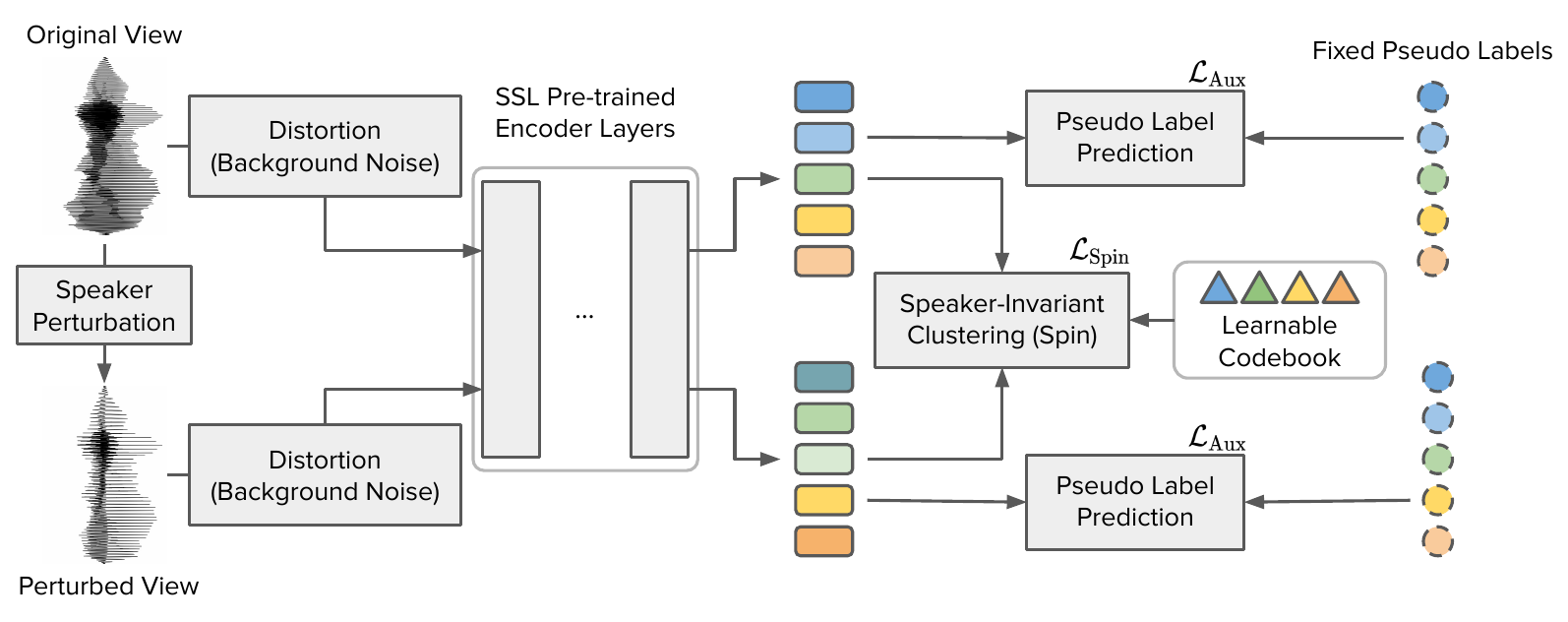}
    \vspace{-10pt}
    \caption{
        The proposed \proposed~domain-specific self-supervision framework.
        The input utterance is perturbed into a different voice and distorted with random noise.
        Both the original and perturbed views are fed into an encoder initialized with an SSL pre-trained model.
        The model is optimized with Speaker-invariant Clustering~(Spin)~\cite{chang2023spin} objective~($\mathcal{L}_{\text{Spin}}$) and frame-wise pseudo-label prediction loss~($\mathcal{L}_{\text{Aux}}$).
    }
    \label{fig:framework}
    \vspace{-6pt}
\end{figure*}

%% file: section/method.tex
\section{Method}
\label{sec:method}

\subsection{Overview}
\label{subsec:method-overview}

The proposed \proposed~framework is shown in Fig.~\ref{fig:framework}.
\proposed~is based on Speaker-invariant Clustering~(Spin)~\citep{chang2023spin}, a domain-specific self-supervision method with online clustering and swapped prediction for capturing content representations~(Sec.~\ref{subsec:method-spin}).
We introduce noise-invariant training by perturbing inputs to improve robustness~(Sec.~\ref{subsec:method-noise}).
Moreover, an auxiliary pseudo-label prediction loss enables fine-tuning the entire model without collapsing~(Sec.~\ref{subsec:method-aux}).
Acoustic Piece is incorporated with the auxiliary loss to improve performance further~(Sec.~\ref{subsec:method-ap}).

\subsection{Speaker-invariant Clustering}
\label{subsec:method-spin}

Spin is an efficient DS method for improving content representations inspired by Swapping Assignments between Views (SwAV)~\citep{caron2020unsupervised}.
We briefly introduce Spin and suggest readers refer to the original paper for further details.

For each utterance in a mini-batch, the F0 frequency and the relative ratio between formant frequencies are randomly perturbed to mimic the same sentence spoken by a different speaker~\citep{choi2021nansy,qian2022contentvec}.
The original and perturbed views are fed into a transformer encoder~\citep{vaswani2017attention} initialized with an SSL model like HuBERT~\citep{hsu2021hubert}.
The output representations $\mathbf{H} = [\boldsymbol{h}_1 \dots \boldsymbol{h}_B]^{\transp}$ of the original view are linearly projected and L2-normalized to $\mathbf{Z} = [\boldsymbol{z}_1 \dots \boldsymbol{z}_B]^{\transp}$, where $B$ is the number of frames in a batch.
We then use the representations to compute a probability distribution over a learnable codebook as $p\left( \cdot | \boldsymbol{z}_b \right)$.
We perform the same operations to the perturbed utterance, resulting in another distribution $q\left( \cdot | \Tilde{\boldsymbol{z}}_b \right)$, where $\Tilde{\cdot}$ denotes features from the speaker-perturbed view.
Next, $q$ is smoothed by solving an optimal transport problem to enforce full codebook usage.
Finally, the model is trained to perform swapped predictions between views by minimizing the cross-entropy loss
\vspace{-4pt}
\begin{equation}
    \begin{aligned}
        \mathcal{L}_{\text{Spin}} = & -\frac{1}{2B} \sum_{b \in [B]} \sum_{k \in [K]} 
        q\left( k | \Tilde{\boldsymbol{z}}_b \right) \log p\left( k | \boldsymbol{z}_b \right) \\[-3pt]
        & -\frac{1}{2B} \sum_{b \in [B]} \sum_{k \in [K]} 
        q\left( k | \boldsymbol{z}_b \right) \log p\left( k | \Tilde{\boldsymbol{z}}_b \right),\\[-3pt]
    \end{aligned}
    \label{eq:spin-loss}
\end{equation}
where $K$ is the size of the learnable codebook, and the second term emerges from the interchangeability of the role of the perturbed and original speech.\footnote{$[N] = \{1, 2, \dots, N \}$ for any positive integer $N$.}
Under this DS framework, the fine-tuned model produces speaker-invariant representations, making the content of speech signals more accessible to downstream applications.

\subsection{Noise-invariant Training}
\label{subsec:method-noise}

To improve the robustness of SSL models, we introduce noise-invariant training by including audio distortions to both views of the input.
We anticipate the model will be able to concurrently eliminate noise and speaker-related information, thereby enabling the trained model to generate robust content representations.

\subsection{Auxiliary Pseudo-label Prediction Loss}
\label{subsec:method-aux}

As noted by \citet{chang2023spin}, Spin is constrained to fine-tuning solely the top layers of pre-trained SSL encoders.
Otherwise, the model converges towards a trivial solution, yielding outputs irrelevant to the corresponding inputs.
This limitation may not be problematic when the application domain closely aligns with the pre-training data.
However, given that the bottom layers are associated with low-level signal processing like denoising~\citep{chang2021end,yuan2023whisperat}, subjecting these layers to fine-tuning is imperative.
This adjustment is particularly beneficial in enhancing the model's robustness to out-of-domain data.
Consequently, we propose a pseudo-label prediction loss to prevent models from collapsing.

The pseudo-label prediction is a frame-wise classification problem with a loss function of
\vspace{-4pt}
\begin{equation}
    \begin{aligned}
        \mathcal{L}_{\text{Aux}} = & -\frac{1}{2B} \sum_{b \in [B]} \log p\left( y_b | \boldsymbol{h}_b \right) \\[-2pt]
        & -\frac{1}{2B} \sum_{b \in [B]} \log p\left( y_b \left| \Tilde{\boldsymbol{h}}_b \right.\right),
    \end{aligned}
    \vspace{-4pt}
    \label{eq:aux-loss}
\end{equation}
where $y_b$ is the pseudo-label at frame $b$.
The probability distributions are computed by projecting $\boldsymbol{h}$ with a fully connected layer followed by a softmax.
The choice of pseudo-labels is flexible, including K-means clusters of acoustic features and codewords produced by Spin.
With this loss, the fine-tuned models are expected to preserve content even when all layers are fine-tuned.
Combining Eqs.~\ref{eq:spin-loss} and \ref{eq:aux-loss}, the overall loss function is
\vspace{-6pt}
\begin{equation}
    \mathcal{L} = \mathcal{L}_{\text{Spin}} + \lambda \mathcal{L}_{\text{Aux}},
    \vspace{-5pt}
    \label{eq:total-loss}
\end{equation}
where $\lambda >$ 0 is a hyper-parameter.
$\mathcal{L}_{\text{Aux}}$ has learning targets independent of the model, regularizing and stabilizing the training process.
Meanwhile, $\mathcal{L}_{\text{Spin}}$ optimizes on varying labels from a dynamically changing codebook, offering flexibility to improve upon the pseudo-labels in $\mathcal{L}_{\text{Aux}}$.
Therefore, the combined loss function is expected to enhance pre-trained speech SSL encoders and mitigate Spin's limitations.

\subsection{Acoustic Piece}
\label{subsec:method-ap}

This section introduces acoustic pieces~\citep{ren2022acoustic-piece} to $\mathcal{L}_{\text{Aux}}$ to further improve \proposed.
APs are learned by applying byte-pair encoding~(BPE)~\citep{sennrich-etal-2016-neural} to discrete acoustic units like K-means clusters of HuBERT representations.
AP captures high-level units close to phonemes and characters, useful for pre-training~\citep{wu2023wav2seq} and generation~\citep{shen2023acoustic}.
Hence, we propose to set AP as the target of $\mathcal{L}_{\text{Aux}}$ to extract better content representations.

Following~\citet{ren2022acoustic-piece}, we first merge identical consecutive units in time for each utterance.
The BPE algorithm is then applied to the reduced sequences to learn acoustic pieces.
Next, we encode the entire training corpus into APs and duplicate the encoded units to the original utterance length.
The encoded corpus is then used as the pseudo-labels for Eq.~\ref{eq:aux-loss}, expecting to encourage the fine-tuned SSL model to encode better phoneme and character representations.

%% file: section/exp.tex
\section{Experiments}
\label{sec:exp}

\subsection{Data}
\label{subsec:exp-data}

The 960 hours of unlabeled English speech in LibriSpeech is used for R-Spin training~\citep{Panayotov2015libri}.\footnote{Released under CC BY 4.0}
Audio distortions are generated with torch-audiomentations.\footnote{https://github.com/asteroid-team/torch-audiomentations}
Following \citet{zhu2023robust}, background noises are sampled from MUSAN~\citep{snyder2015musan} and CHiME-4~\citep{vincent2017chime4}, covering music, speech, and outdoor noise.\footnote{MUSAN: CC BY 4.0 / CHiME-4: CC BY-NC-SA 2.0}
Signal-to-noise ratios~(SNR) are uniformly sampled from [$-$10, 10] during training.
We add distortions to each utterance during evaluation, including Gaussian noise, MUSAN noise, and reverberation~(Appendix~\ref{subsec:app-impl-pr}).

\subsection{Implementation}
\label{subsec:exp-impl}
The DS experiments are mostly based on WavLM~\citep{chen2022wavlm} because WavLM is pre-trained with a denoising objective, offering a good initialization.
HuBERT~\citep{hsu2021hubert} is also considered to demonstrate R-Spin's generalizability to SSL models trained with clean speech.
We follow the implementations by~\citet{chang2023spin}, which uses PyTorch~\citep{paszke2019pytorch}, PyTorch-Lightning~\citep{falcon2019pytorch-lightning}, and torchaudio~\citep{yang2022torchaudio}.\footnote{https://github.com/vectominist/spin}
The acoustic pieces are generated by learning BPE tokens on top of a HuBERT + Spin\textsubscript{2048} model~(Appendix~\ref{subsec:app-impl-spin}).
Further details can be found in Appendix~\ref{subsec:app-impl-rspin}.\footnote{We employ GitHub Copilot for implementation assistance and ChatGPT for writing refinement.}

\subsection{Notations}
\label{subsec:notation}

We denote an SSL model $X$ fine-tuned with Spin and $K$ codewords with $X$ + Spin\textsubscript{$K$}.
In $X$ + R-Spin\textsubscript{$K_1, K_2$}, $K_1$ and $K_2$ are respectively the codebook size of $\mathcal{L}_{\text{Spin}}$ and the number of classes of pseudo-labels for $\mathcal{L}_{\text{Aux}}$.
If the pseudo-labels are acoustic pieces, ``AP'' is added to $K_2$.
Unless specified otherwise, R-Spin denotes R-Spin\textsubscript{32, AP40k}.

\input{table/pr_main}

\subsection{Noisy Phoneme Recognition}
\label{subsec:exp-noisy-pr}

We compare the phoneme recognition performance of SSL and DS methods under noisy conditions.
The training setup is similar to the SUPERB phoneme recognition task~\citep{yang2021superb}, where the SSL models are frozen and only a lightweight prediction head is fine-tuned~(Appendix~\ref{subsec:app-impl-pr}).\footnote{https://github.com/s3prl/s3prl}
We apply some classes of distortions only to testing data to obtain phoneme error rates~(PER).
We divide results by budget, which is the amount of speech processed during DS, proportional to the computational resources required~(Sec.~\ref{subsec:exp-proc-data}).

As shown in the middle columns of Table~\ref{tab:pr-asr}, R-Spin outperforms low and high-budget methods in all conditions.
WavLM + R-Spin has the best overall PERs because WavLM is pre-trained with a denoising task, showing that model initialization contributes largely to the recognition performance after DS.
Next, R-Spin improves unseen tasks like Gaussian noise and reverberation, indicating that noise-invariant training generalizes to some out-of-domain perturbations.
Furthermore, comparing Robust data2vec with R-Spin is unfair since the training costs are 12 times apart, so we train a low-budget Robust data2vec~(Appendix~\ref{subsec:app-impl-rd2v}).
The noticeable performance drop in the low-budget model implies Robust data2vec requires high computation resources, but our approach offers competitive results with fewer training data.\footnote{
    The low-budget version reduces the number of GPUs to match the amount of training data processed with R-Spin, but the hyperparameters are difficult to tune, leading to significantly degraded performance.
}

\input{fig/snr}

We plot PERs under different SNRs in Fig.~\ref{fig:snr-pr} for a detailed comparison.
Overall, R-Spin achieves the lowest PERs even when the SNR is high.
HuBERT-MGR excels in Gaussian noise because it is the only model trained with this noise type.
Nevertheless, R-Spin offers a similar performance in Gaussian noise across different SNRs, aligning with the results in Table~\ref{tab:pr-asr}.

\subsection{Noisy Speech Recognition}
\label{subsec:exp-noisy-asr}

This section assesses R-Spin with a noisy ASR task.
We adopt the SUPERB ASR setup with the CHiME-4 corpus~\citep{vincent2017chime4} to evaluate the models in more realistic noisy recordings~(Appendix~\ref{subsec:app-impl-asr}).
The results in the right columns of Table~\ref{tab:pr-asr} reveal that R-Spin surpasses low-budget baseline models.
While R-Spin demonstrates commendable performance on CHiME-4, this method falls short compared to Robust data2vec, which benefits from training with a substantially higher budget.
Furthermore, we set Whisper Base and Small as toplines due to their robustness demonstrated through large-scale weakly-supervised learning~\citep{radford2022whisper}.
R-Spin successfully mitigates the performance gap between WavLM and the Whisper toplines by over 60\%.
Combining phoneme and speech recognition findings, we conclude that R-Spin effectively enhances pre-trained SSL models in capturing robust content representations.

\subsection{Data Efficiency}
\label{subsec:exp-proc-data}

Developing R-Spin aims to enhance speech SSL models with minimal resources, including improving data efficiency.
Following \citet{chang2023spin}, an analysis of the duration of speech data processed during training is undertaken to quantify the computational expenses associated with each method.
As depicted in the second column of Table~\ref{tab:pr-asr}, these values are derived by multiplying the number of training updates and the effective batch size for each update.
Compared with the high-budget methods, R-Spin requires significantly lower training costs, concurrently exhibiting superior performance across diverse conditions.
A complete comparison of the costs can be found in Appendix~\ref{sec:app-compute-resource}.

\input{fig/trajectory}

\input{fig/cka}

\input{fig/sid}

\subsection{Representation Invariability}
\label{subsec:exp-inv}

This section explores the robustness of models regarding representation invariability by examining their characteristics under diverse perturbations.

\subsubsection{Speaker Invariability}
\label{subsubsec:exp-spk-inv}

We first inspect each model's invariability to speaker changes by computing the speaker identification~(SID) accuracy with different hidden layer representations.
The SID task follows SUPERB's setup but with 50k training updates.
As shown in Fig.~\ref{fig:sid}, \proposed~has a significantly lower SID accuracy for the top layers, demonstrating the effect of fine-tuning the whole model with a speaker-invariant objective.
Moreover, requiring 9X less training costs, our method produces representations with less speaker information than ContentVec.
Therefore, the proposed method outperforms prior speaker-invariant self-supervision approaches in removing speaker ID.

Next, we use t-SNE~\citep{van2008tsne} to visualize representations articulated by distinct speakers.
We show the layer with the lowest SID rate according to Fig.~\ref{fig:sid}.
In Figs.~\ref{fig:traj-hubert-L0} and \ref{fig:traj-rspin-L0}, there is a discernible clustering of frames uttered by the same speaker, suggesting that lower layers retain more speaker-specific information.
Conversely, Figs.~\ref{fig:traj-hubert-L9} and \ref{fig:traj-rspin-L12} illustrate that top layer features are grouped according to phonemes rather than speakers.
Moreover, the top layer representations are context-dependent, as exemplified by the spatial arrangement of phonemes such as ``carry'' (\texttt{k eh r iy}) and the same phoneme \texttt{/iy/} in the word "oily" (\texttt{oy l iy}).
Besides, a comparative analysis between Figs.~\ref{fig:traj-hubert-L9} and \ref{fig:traj-rspin-L12} reveals that R-Spin features exhibit a more prominent overlap among speakers than HuBERT.
As a result, this section substantiates the speaker-invariability of R-Spin.
Detailed visualization can be found in Appendix~\ref{sec:app-tsne-repr}.

\input{fig/analysis_figs}

\subsubsection{Noise Invariability}
\label{subsubsec:exp-noise-inv}
\input{fig/trajectory_domain}

We examine the response of continuous representations to input distortions.
We compute linear centered kernel alignment~(CKA) similarities~\citep{kornblith2019cka} of frame-wise features with and without noisy inputs, where a higher similarity indicates a higher invariability to distortions.
The evaluation involves building datasets derived from the LibriSpeech dev-clean and dev-other sets augmented with various distortions.
Fig.~\ref{fig:cka} illustrates that R-Spin exhibits superior noise invariability for the upper layers than other models, indicating the efficacy of noise-invariant training even if the noise types are unseen.
Lower layers tend to have lower similarities, suggesting a higher sensitivity to perturbations.
This observation aligns with existing research discussed in Sec.~\ref{subsec:method-aux}, which associates lower layers with fundamental signal processing functions.
In contrast, Robust data2vec has a greater noise invariability starting from the bottom layers because of the data2vec training strategy.

We employ t-SNE to explore representations under distortions.
As shown in Fig.~\ref{fig:trajectory-domain}, the R-Spin features exhibit a more pronounced overlap than HuBERT, suggesting that R-Spin improves robustness to noise, aligning with the observations in Fig.~\ref{fig:cka}.
Fig.~\ref{fig:traj-rspin-dom} reveals that R-Spin features exposed to MUSAN noise exhibit a high degree of overlap with unperturbed ones, whereas the other two perturbation types diverge slightly because Gaussian noise and reverberation are unseen for R-Spin.
Overall, the analysis underscores the notable noise invariability offered by R-Spin.

\subsection{Importance of Discrete Units}
\label{subsec:exp-discrete-units}

This section analyzes the efficacy of APs and their relation to phonemes and characters.

\subsubsection{Codebook and Acoustic Pieces Size}
\label{subsubsec:exp-codebook-ap}

We inspect the importance of the codebook size in Spin.
As highlighted by \citet{chang2023spin}, the codebook size positively correlates with phoneme recognition.
A similar trend can be found in Fig.~\ref{fig:k-per-wer} but has an inverted trend for ASR.
However, the observed performance discrepancy is less than 1\% absolute, suggesting that codebook size's impact on R-Spin is marginal.
In contrast, substantial improvements in ASR are observed with more APs, but not in phoneme recognition, as evidenced by Figs.~\ref{fig:bpe-word} and \ref{fig:bpe-phone}.
To analyze this phenomenon, we investigate R-Spin's phoneme and character segmentation capabilities using discrete units.

\input{fig/phone_align_example}

\subsubsection{Phoneme and Character Segmentation}
\label{subsubsec:exp-phone-seg}

We segment speech with acoustic pieces and show the R-values in Figs.~\ref{fig:bpe-phone} and~\ref{fig:bpe-word}.
R-value, a metric for evaluating word or phoneme segmentation quality~\citep{rasanen2009rval}, is robust to over-segmentation, an issue that plagues F1.
The boundaries are predicted by locating differing adjacent discrete units.
We evaluate on force-aligned LibriSpeech dev-clean and dev-other sets~\citep{lugosch2019speech,mcauliffe2017montreal}.\footnote{https://zenodo.org/record/2619474 (CC-BY 4.0)}
The character boundaries are obtained by dividing each force-aligned word segment into equal-length segments corresponding to individual characters within the word.
More accurate boundaries can be computed with character-based aligners, but we only need a rough estimation of the segmentation quality.

As depicted in both Figs.~\ref{fig:bpe-phone} and \ref{fig:bpe-word}, larger AP vocabulary sizes have superior segmentation, indicating that more APs form units that closely resemble linguistic units.
The baseline, which involves uniformly segmenting utterances based on the number of boundaries derived from APs, underscores the non-random nature of AP boundaries.
Although the segmentation capability of APs is incomparable with other unsupervised speech segmentation algorithms~\citep{kreuk2020self}, they present significantly improved targets for $\mathcal{L}_{\text{Aux}}$, consequently enhancing the accuracy of ASR.
A full comparison of unsupervised phoneme segmentation can be found in Appendix~\ref{subsec:app-segment}.

Furthermore, we provide an example of segmenting an utterance with 40k APs in Fig.~\ref{fig:phone-align}.
The red dashed stripes depict that the boundaries of APs are mostly aligned with phoneme boundaries.
Notably, the predicted boundaries occasionally exhibit a slight temporal lag compared to the ground truth, like the first \texttt{/ah/} and \texttt{/m/}.
We suspect the 50Hz framerate of HuBERT or the Spin training objective causes this phenomenon since they could reduce time resolution and introduce temporal shifts.
Still, the actual cause remains a subject for future investigation.
To summarize, APs effectively learn discrete acoustic units that benefit ASR performance.

\subsection{Ablation Studies}
\label{subsec:exp-ablation}

\input{table/ablation}

Under the same CHiME-4 ASR setup in Sec.~\ref{subsec:exp-noisy-asr}, we conduct ablation studies to analyze the proposed methods.
As shown in Table~\ref{tab:ablation}, the WERs increase significantly without $\mathcal{L}_{\text{Aux}}$), showing that the auxiliary loss helps ASR performance and mitigates collapsing.
Second, WERs increase by about 5\% without $\mathcal{L}_{\text{Spin}}$, indicating the necessity of this loss for achieving perturbation-invariant representations.
Speaker perturbation also plays an important role in offering good content representations according to the degraded WERs.
Moreover, the fine-tuned model exhibited suboptimal performance when trained without noise, emphasizing the importance of noise-invariant training for improving robustness.
The above findings verify the necessity of the proposed approaches.

We inspect the effect of choosing different pseudo-labels for $\mathcal{L}_{\text{Aux}}$.
First, APs are helpful for R-Spin since learning from the original Spin model's codeword labels increases WERs by over 2\%.
Next, we replace the pseudo-labels with the more commonly used K-means clustered representations~\citep{hsu2021hubert}.
Clustered MFCC features degrade R-Spin the most, no matter the number of clusters used.
In contrast, clustered HuBERT representations from layer 9~(L9) yield results comparable to Spin\textsubscript{2048}, and the t-test suggests the disparities between pseudo-labels are statistically insignificant.
Thus, using clustered features from an SSL model is a viable alternative.
Additional ablation studies are available in Appendix~\ref{subsec:app-ablation}.

%% file: table/pr_main.tex
\begin{table*}[t]
    \centering
    \begin{adjustbox}{max width=0.88\linewidth}
    \begin{threeparttable}
    \begin{tabular}{@{~}l@{~}c@{~~}c@{~~}c@{~~}c@{~~}c@{~~}c@{~~}c@{~~}c@{~~}c@{~}}
        \toprule
        & \multirow{3}{*}{\shortstack[c]{~\\Processed\\ Speech \\(hours)}} & \multicolumn{4}{@{~~}c@{~~}}{\multirow{2}{*}{\shortstack[c]{LibriSpeech test-other\\Phoneme Recognition (PER$\downarrow$)}}} & & \multicolumn{2}{@{~~}c@{~}}{\multirow{2}{*}{\shortstack[c]{CHiME-4\\ASR (WER$\downarrow$)}}} \\
        \\
        \cmidrule{3-6}
        \cmidrule{8-9}
        \cmidrule{4-6}
        Method & & Clean & Gaussian$^{\dagger}$ & \textsc{Musan} & Reverb$^{\dagger}$ & & Real & Sim \\
        \midrule
        \midrule
        \multicolumn{3}{@{~}l@{~~}}{\textbf{No DS Baselines}} \\
        ~~HuBERT~\citep{hsu2021hubert} & 0 & 10.7 & 74.5 & 50.2 & 23.2 & & 72.7 & 63.1 \\
        ~~WavLM~\citep{chen2022wavlm} & 0 & 10.3 & 59.9 & 45.1 & 19.4 & & 52.4 & 46.4 \\

        \midrule
        \multicolumn{3}{@{~}l@{~~}}{\textbf{DS Baselines}} \\
        ~~HuBERT + Spin\textsubscript{2048}~\cite{chang2023spin} & 0.4k & 8.4 & 70.8 & 47.8 & 18.4 & & 71.3 & 62.0 \\
        ~~WavLM + Spin\textsubscript{2048}~\cite{chang2023spin} & 0.4k & \textbf{8.2} & 59.2 & 41.2 & 16.7 & & 52.1 & 46.6 \\
        ~~Robust data2vec (Low-budget) & 10.4k & 38.8 & 68.2 & 52.9 & 53.7 & & 80.9 & 78.2 \\

        \midrule
        \multicolumn{3}{@{~}l@{~~}}{\textbf{Proposed}} \\
        ~~HuBERT + R-Spin\textsubscript{32, AP40k} & 8.2k & 8.3 & 36.4 & 18.2 & 16.3 & & 34.3 & 34.1 \\
        ~~WavLM + R-Spin\textsubscript{32, AP40k} & 8.2k & \textbf{8.2} & \textbf{33.7} & \textbf{16.7} & \textbf{14.9} & & \textbf{26.4} & \textbf{26.6} \\

        \midrule
        \midrule
        \multicolumn{3}{@{~}l@{~~}}{\textbf{High-budget DS Toplines}} \\
        ~~ContentVec\textsubscript{500}~\citep{qian2022contentvec} & 76k & 8.7 & 71.4 & 47.2 & 16.8 & & 61.4 & 55.1 \\
        ~~HuBERT-MGR~\citep{huang2022improving} & 78k & 9.5 & 37.1 & 36.3 & 18.3 & & 49.7 & 44.3 \\
        ~~Robust data2vec~\citep{zhu2023robust} & 105k & 6.5 & 56.7 & 27.7 & 19.2 & & 17.5 & 20.1 \\

        \midrule
        \multicolumn{3}{@{~}l@{~~}}{\textbf{Supervised Toplines}} \\
        ~~Whisper Base~\citep{radford2022whisper} & -- & -- & -- & -- & -- & & 17.9 & 23.3 \\
        ~~Whisper Small~\citep{radford2022whisper} & -- & -- & -- & -- & -- & & 10.8 & 14.3 \\
        \bottomrule
    \end{tabular}
    \begin{tablenotes}[flushleft]
        \item 
            $^{\dagger}${\small Unseen perturbation types for \proposed~and Robust data2vec.}
    \end{tablenotes}
    \end{threeparttable}
    \end{adjustbox}
    \vspace{-6pt}
    \caption{
        Phoneme recognition on LibriSpeech and ASR on CHiME-4 test sets.
        Gaussian noise, MUSAN background noise, and reverberation~(Reverb) are respectively added to simulate noisy conditions, where the SNRs are fixed to 0dB.
        The calculation of the number of hours of processed speech during DS follows~\citet{chang2023spin}.
    }
    \label{tab:pr-asr}
    \vspace{-5pt}
\end{table*}

%% file: fig/snr.tex
\begin{figure}
    \centering
    \begin{subfigure}[b]{\linewidth}
        \centering
        \includegraphics[width=0.9\linewidth]{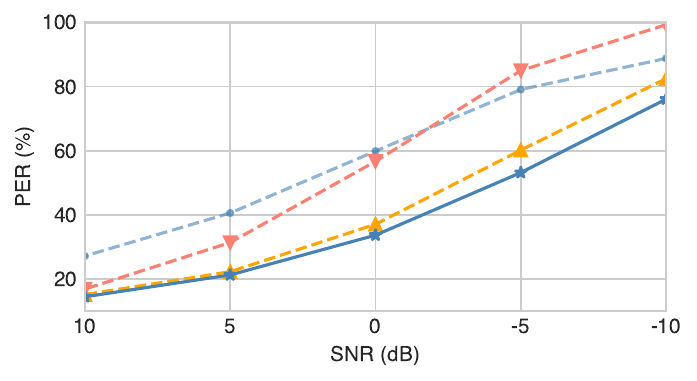}
        \vspace{-6pt}
        \caption{Gaussian Noise}
        \label{fig:snr-pr-colored}
    \end{subfigure}
    \begin{subfigure}[b]{\linewidth}
        \centering
        \includegraphics[width=0.9\linewidth]{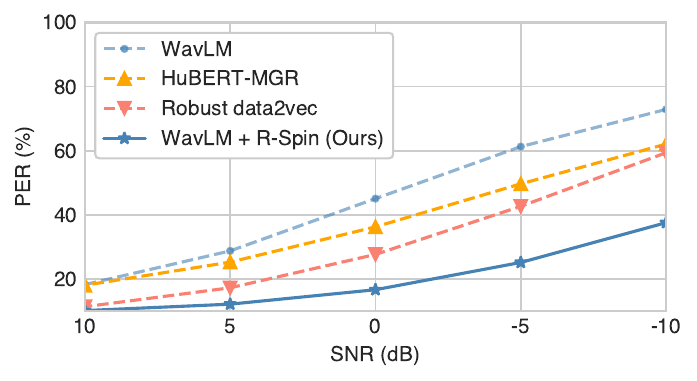}
        \vspace{-6pt}
        \caption{MUSAN Noise}
        \label{fig:snr-pr-musan}
    \end{subfigure}
    \vspace{-20pt}
    \caption{
        Phoneme error rates~(PER) under different noise types and SNRs.
        \proposed\textsubscript{32, AP40k} is used here.
    }
    \label{fig:snr-pr}
    \vspace{-6pt}
\end{figure}

%% file: fig/trajectory.tex
\begin{figure*}
    \centering
    \begin{subfigure}[b]{0.2455\linewidth}
        \centering
        \captionsetup{justification=centering}
        \includegraphics[width=1.15\linewidth]{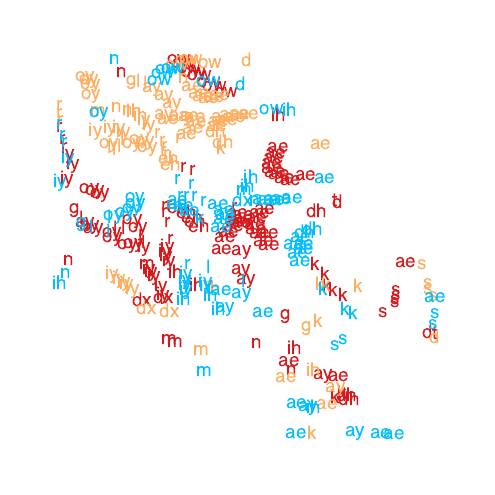}
        \vspace{-27pt}
        \caption{HuBERT\\CNN (Layer 0)}
        \label{fig:traj-hubert-L0}
    \end{subfigure}
    \begin{subfigure}[b]{0.2455\linewidth}
        \centering
        \captionsetup{justification=centering}
        \includegraphics[width=1.15\linewidth]{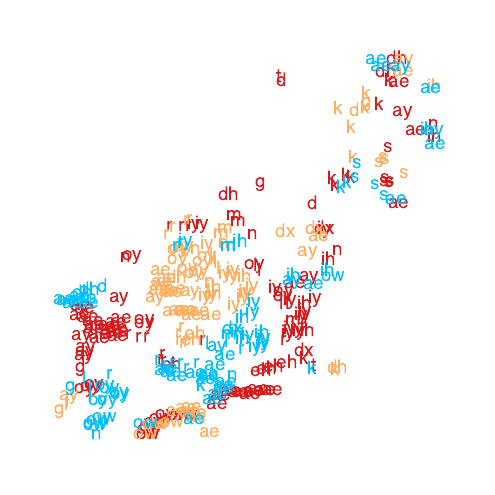}
        \vspace{-27pt}
        \caption{HuBERT + R-Spin\\CNN (Layer 0)}
        \label{fig:traj-rspin-L0}
    \end{subfigure}
    \begin{subfigure}[b]{0.2455\linewidth}
        \centering
        \captionsetup{justification=centering}
        \includegraphics[width=1.15\linewidth]{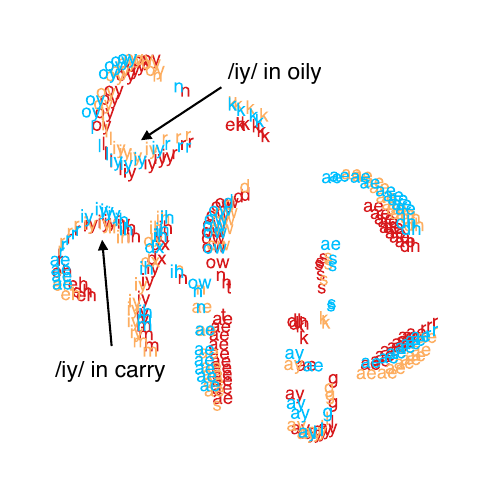}
        \vspace{-27pt}
        \caption{HuBERT\\Best Layer (9)}
        \label{fig:traj-hubert-L9}
    \end{subfigure}
    \begin{subfigure}[b]{0.2455\linewidth}
        \centering
        \captionsetup{justification=centering}
        \includegraphics[width=1.15\linewidth]{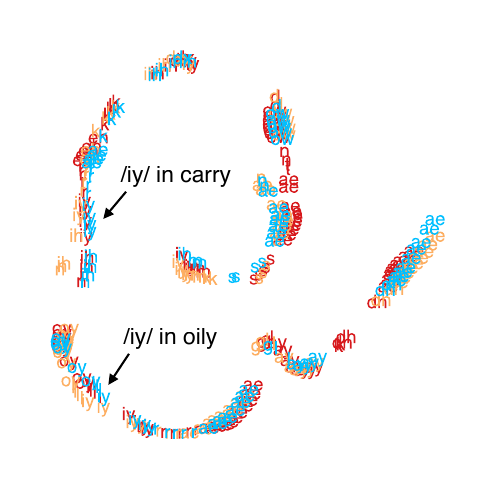}
        \vspace{-27pt}
        \caption{HuBERT + R-Spin\\Best Layer (12)}
        \label{fig:traj-rspin-L12}
    \end{subfigure}
    \vspace{-20pt}
    \caption{
        t-SNE~\citep{van2008tsne} visualization of the CNN and the layer with the lowest speaker identification rate given the same clean utterance spoken by three speakers from TIMIT~\citep{garofolo1993timit}.
        Each color represents a speaker, while each label visualizes a frame and the corresponding phoneme label.
        The transcription is ``Don't ask me to carry an oily rag like that.''
        The silence frames are omitted for clarity.
    }
    \label{fig:trajectory}
    \vspace{-8pt}
\end{figure*}


%% file: fig/cka.tex
\begin{figure*}
    \centering
     \begin{subfigure}[b]{0.485\linewidth}
         \centering
         \includegraphics[width=\linewidth]{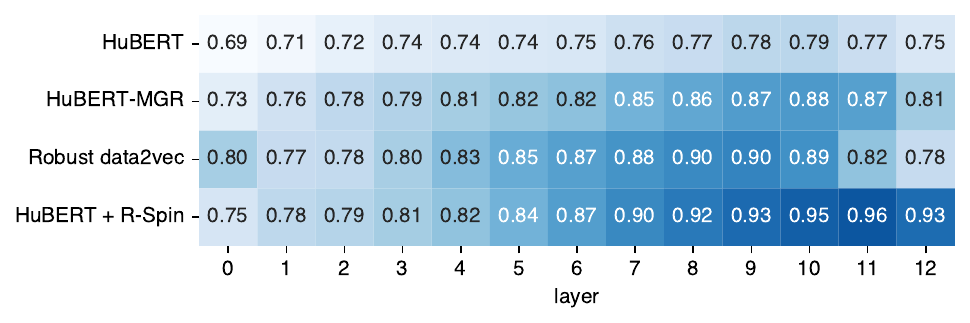}
         \vspace{-18pt}
         \caption{Gaussian Noise (SNR $=$ 0dB)}
         \label{fig:cka-colored}
     \end{subfigure}
     \begin{subfigure}[b]{0.485\linewidth}
         \centering
         \includegraphics[width=\linewidth]{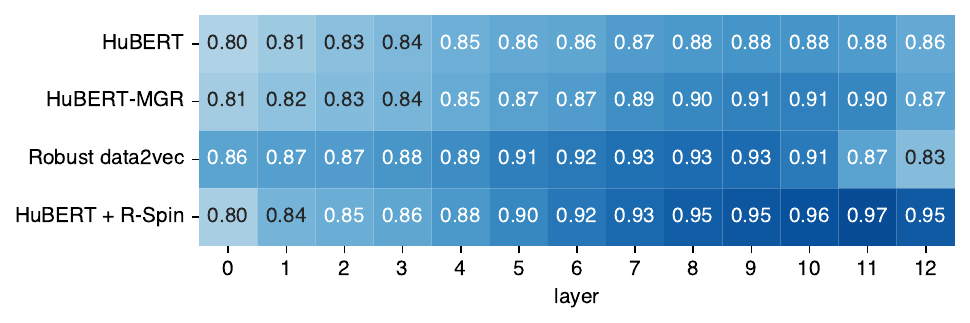}
         \vspace{-18pt}
         \caption{Reverberation (real room impulse response)}
         \label{fig:cka-reverb-real}
     \end{subfigure}
     \vspace{-6pt}
    \caption{
        Layer-wise perturbation invariability analyses with Linear CKA, where higher values indicate higher invariability to perturbations.
        The zeroth layer denotes the CNN feature extractor.
    }
    \label{fig:cka}
    \vspace{-10pt}
\end{figure*}

%% file: fig/sid.tex
\begin{figure}
    \centering
    \includegraphics[width=0.85\linewidth]{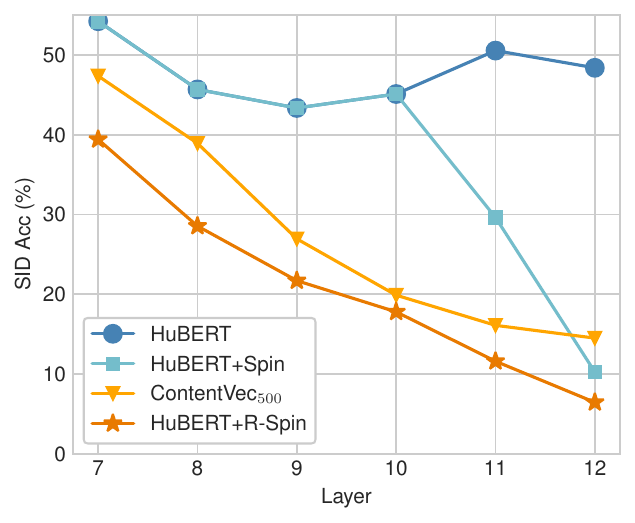}
    \vspace{-10pt}
    \caption{Layer-wise speaker identification accuracy.}
    \label{fig:sid}
    \vspace{-10pt}
\end{figure}

%% file: fig/analysis_figs.tex
\begin{figure*}
    \centering
    \begin{subfigure}[b]{0.329\linewidth}
         \centering
         \includegraphics[width=1.03\linewidth]{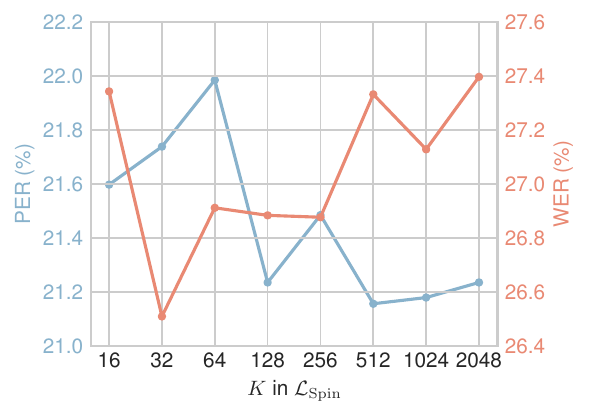}
         \vspace{-20pt}
         \caption{}
         \label{fig:k-per-wer}
     \end{subfigure}
     \begin{subfigure}[b]{0.329\linewidth}
         \centering
         \includegraphics[width=\linewidth]{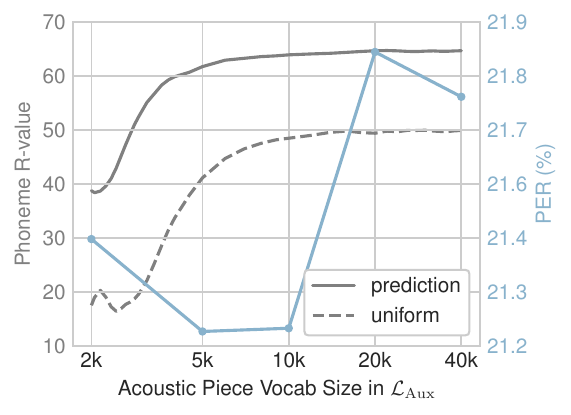}
         \vspace{-20pt}
         \caption{}
         \label{fig:bpe-phone}
     \end{subfigure}
     \begin{subfigure}[b]{0.329\linewidth}
         \centering
         \includegraphics[width=\linewidth]{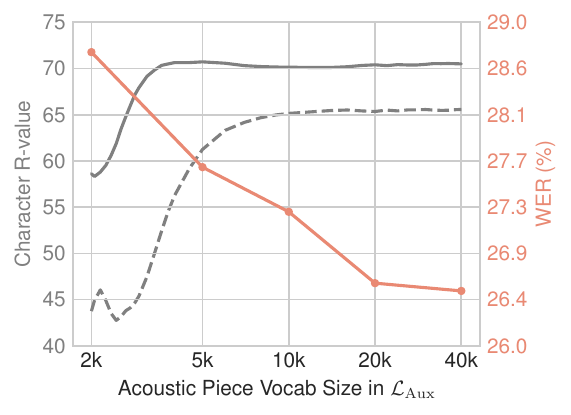}
         \vspace{-20pt}
         \caption{}
         \label{fig:bpe-word}
     \end{subfigure}
     \vspace{-20pt}
    \caption{
        WavLM + \proposed~with different (a) codebook and (b)(c) AP vocabulary sizes.
        (b) and (c) depict the phoneme and character segmentation R-values, where the dotted curves are the baselines by segmenting each utterance with equal-length segments given the number of boundaries obtained by the APs.
        The PERs are calculated by averaging over different noise conditions on LibriSpeech test-other.
        The WERs are the averaged scores of the real and simulated evaluation sets of CHiME-4.
    }
    \label{fig:analysis-figs}
    \vspace{-10pt}
\end{figure*}

%% file: fig/trajectory_domain.tex
\begin{figure}
    \centering
    \begin{subfigure}[b]{0.48\linewidth}
        \centering
        \captionsetup{justification=centering}
        \includegraphics[width=1.08\linewidth]{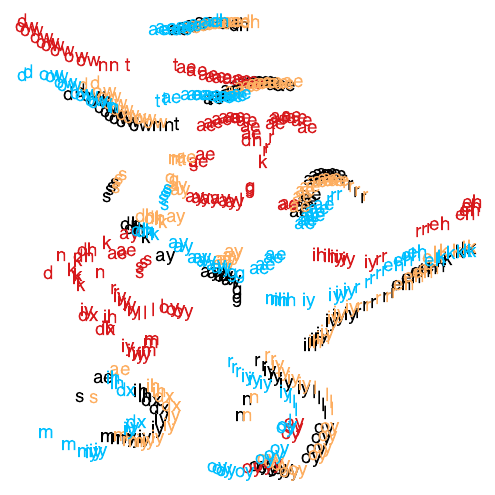}
        \vspace{-15pt}
        \caption{HuBERT\\Best Layer (9)}
        \label{fig:traj-hubert-dom}
    \end{subfigure}
    \begin{subfigure}[b]{0.48\linewidth}
        \centering
        \captionsetup{justification=centering}
        \includegraphics[width=1.08\linewidth]{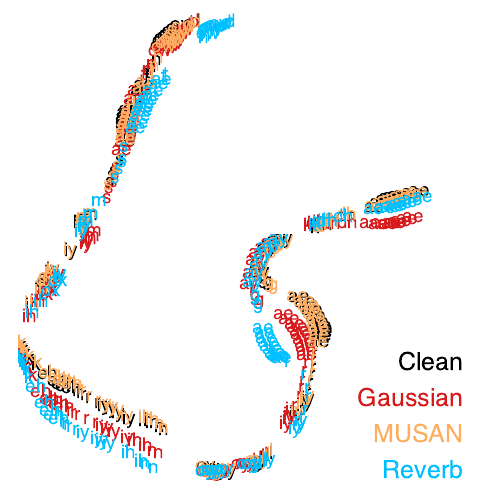}
        \vspace{-15pt}
        \caption{HuBERT~+~R-Spin\\Best Layer (12)}
        \label{fig:traj-rspin-dom}
    \end{subfigure}
    \vspace{-6pt}
    \caption{
        t-SNE visualization of hidden representations of the same utterance in Fig.~\ref{fig:trajectory} with different distortions indicated by colors, where SNR $=$ 0dB.
    }
    \label{fig:trajectory-domain}
    \vspace{-5pt}
\end{figure}


%% file: fig/phone_align_example.tex
\begin{figure*}[t]
    \centering
    \includegraphics[width=\linewidth]{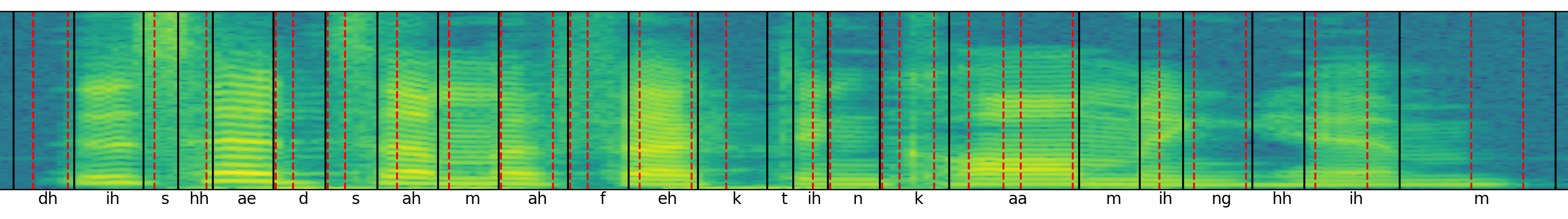}
    \vspace{-24pt}
    \caption{
        An example of phoneme alignment of an utterance ``This had some effect in calming him.'' from LibriSpeech dev-clean.
        The black lines indicate the force-aligned boundaries, while the red dashed lines are the predicted boundaries with AP40k.
    }
    \label{fig:phone-align}
    \vspace{-10pt}
\end{figure*}

%% file: table/ablation.tex
\begin{table}[t]
    \centering
    \begin{adjustbox}{max width=0.85\linewidth}
    \begin{threeparttable}
    \begin{tabular}{lcc}
        \toprule
        & \multicolumn{2}{c}{CHiME-4} \\
        Method & Real & Sim \\
        \midrule
        Spin\textsubscript{2048}~\citep{chang2023spin} & 52.1 & 46.6 \\
        \proposed\textsubscript{32, AP40k} (Proposed) & \textbf{26.4} & \textbf{26.6} \\
        ~~no $\mathcal{L}_{\text{Aux}}$ & 47.8 & 45.6 \\
        ~~no $\mathcal{L}_{\text{Spin}}$ & 31.9 & 32.4 \\
        ~~no speaker perturbation & 28.3 & 28.0 \\
        ~~no additive noise & 49.4 & 46.8 \\
        \midrule
        \multicolumn{3}{l}{\textbf{Pseudo Label for} $\mathcal{L}_{\text{Aux}}$} \\
        ~~Spin\textsubscript{2048} codebook$^{\spadesuit}$ & 28.3 & 29.1 \\
        ~~MFCC (K-means 512) & 46.9 & 45.4 \\
        ~~MFCC (K-means 2048) & 48.5 & 45.5 \\
        ~~HuBERT L9 (K-means 512)$^{\spadesuit}$ & 28.8 & 29.1 \\
        ~~HuBERT L9 (K-means 2048)$^{\spadesuit}$ & 28.2 & 28.4 \\
        \bottomrule
    \end{tabular}
    \begin{tablenotes}[flushleft]
        \item 
            $^{\spadesuit}${\small Pairwise t-tests between these results all have $p >$ 0.05. Also, $p <$ 0.05 when they are compared with \proposed\textsubscript{32, AP40k}.}
    \end{tablenotes}
    \end{threeparttable}
    \end{adjustbox}
    \vspace{-5pt}
    \caption{CHiME-4 ASR results for ablation studies based on fine-tuned WavLM models.}
    \label{tab:ablation}
    \vspace{-10pt}
\end{table}

%% file: section/conclusion.tex
\section{Conclusion}
\label{sec:conclusion}

This paper proposes R-Spin, a domain-specific self-supervision method with speaker and noise-invariant clustering for robust content representations.
Results illustrate the efficacy and broad applicability of R-Spin across various acoustic scenarios, even within constrained computation budgets.
The acoustic analyses presented in this study offer insights into the characteristics of discrete units of this nature and strategies for their utilization.
Future directions involve scaling to larger models and exploring its application in diverse downstream tasks like robust voice conversion.

%% file: section/ack.tex
\section*{Acknowledgements}
\label{sec:ack}

We thank Alexander H. Liu, Saurabhchand Bhati, Nauman Dawalatabad, and Yuan Gong for their insightful feedback.

%% file: section/limitation.tex
\section*{Limitations}
\label{sec:limitation}

This paper faces four primary limitations due to constrained computation resources and available data.
First, we consider background noises including human speech, music, and natural noises, and evaluate the proposed methods with similar noise types and reverberation, covering many real-world conditions.
However, the trained models may encounter challenges in processing more severely distorted audio data, such as air traffic control communications.
Second, the dataset employed consists of English utterances spoken by native speakers, predominantly of North American dialects, leaving the performance in other languages and accents unexplored.
Third, the experiments are conducted on 95M-parameter models, so the scalability of R-Spin remains unknown.
Last, to fully comprehend the capabilities of the proposed method, further analyses and extensions to other applications are recommended for future exploration~\citep{sicherman2023analysing}.
These questions can be answered by experimenting with diverse datasets and more computation resources.

%% file: section/ethics.tex
\section*{Ethics Statement}
\label{sec:ethics}

Our models inherit the biases of SSL models (HuBERT and WavLM) pre-trained on the LibriSpeech corpus.
This corpus contains read English audio recordings derived from audiobooks.
Limitations arise when confronted with accents and topic domains outside the corpus scope, potentially diminishing the effectiveness of the proposed methods.
Thus, the direct application of our models to real-world scenarios may result in increased speech recognition error rates.
These errors, if unaddressed, can propagate through downstream applications like natural language processing systems, leading to potential risks for users, such as the misinterpretation of voice commands.

%% file: section/appendix.tex
\appendix

\section{Implementation Details}
\label{sec:app-impl-detail}

\subsection{Speech SSL Models}
\label{subsec:app-impl-ssl}

Each SSL model used in this paper has a 7-layer CNN feature extractor and a 12-layer transformer encoder~\citep{vaswani2017attention}, having roughly 95M parameters in total.
All SSL models are pre-trained with 960 hours of unlabeled speech in the LibriSpeech corpus.
\textbf{HuBERT}~\citep{hsu2021hubert} is pre-trained in two iterations.
In the first iteration, the encoder model learns to predict each masked frame's K-means cluster ID of MFCC features.
The second iteration model has learning targets obtained by clustering hidden representations of the first iteration model.
\textbf{WavLM}~\citep{chen2022wavlm} follows the second iteration of HuBERT, but the training process involves a denoising task by adding random noises to the input to increase robustness.
\textbf{ContentVec}~\citep{qian2022contentvec} is fine-tuned on top of a pre-trained HuBERT model, but the inputs are augmented with speaker perturbation so that the model learns to produce representations invariant of the speaker.
ContentVec's learning targets are obtained by converting all LibriSpeech data into the same speaker with a voice conversion model and then applying K-means clustering to the hidden features of HuBERT, given the converted inputs.
\textbf{HuBERT-MGR}~\citep{huang2022improving} continues the HuBERT pre-training process with noisy speech and an auxiliary domain adversarial training objective to enhance robustness.
HuBERT-MGR is trained with a mix of clean and distorted speech, where the distortions include MUSAN background noise, Gaussian noise, and reverberation.
\textbf{Robust data2vec}~\citep{zhu2023robust} fine-tunes a pre-trained data2vec model.
Unlike data2vec, the inputs to the student model include background noise so that the model learns denoising.
An additional contrastive learning objective is incorporated to enhance robustness.
The pre-trained model weights are obtained from the s3prl toolkit.\footnote{https://github.com/s3prl/s3prl/tree/main/s3prl/upstream}

\subsection{Spin}
\label{subsec:app-impl-spin}

Since \proposed~is trained with 960 hours of speech in LibriSpeech, the pseudo-labels for $\mathcal{L}_{\text{Aux}}$ should be generated for all those data with Spin.
To avoid labeling unseen data with Spin, we train another HuBERT + Spin\textsubscript{2048} model with the same data~(originally 100 hours in \citealp{chang2023spin}).
Each mini-batch before data perturbation has 2560 seconds of speech, equivalent to 32k frames after downsampling.
The learning rate first linearly increases from 10\textsuperscript{$-$6} to 10\textsuperscript{$-$4} in the first 2.5k updates, then linearly decreases to 10\textsuperscript{$-$6} in the last 7.5k updates.
The implementation of the Spin loss follows \citet{caron2020unsupervised}.\footnote{https://github.com/facebookresearch/swav}
This model takes four hours of training time on four RTX A6000 GPUs.
Models trained with all 10k updates are used to generate pseudo-labels.
In total, roughly 7.1k hours of unlabeled speech data are processed.
Compared with the model in \citet{chang2023spin}, a similar performance is achieved on phoneme recognition.

\subsection{R-Spin}
\label{subsec:app-impl-rspin}

Each mini-batch before perturbation has 384 seconds of speech, equivalent to 19.2k frames in each view.
Each utterance is first speaker-perturbed to generate a second view.
All utterances are added with noise from MUSAN and CHiME-4 with an SNR in [$-$10, 10] dB.
The noise for each view is independent.
The learning rate first linearly increases from 10\textsuperscript{$-$6} to 10\textsuperscript{$-$4} in the first 4k updates, then linearly decreases to 10\textsuperscript{$-$6} in the last 6k updates.
$\lambda$ in Eq.~\ref{eq:total-loss} is set to 5.
Each \proposed~DS training takes less than eight hours on two RTX A6000 GPUs.
Models trained with all 10k updates are used for evaluation.
For the \proposed~training, 1.1k hours of unlabeled speech data are processed.
Combined with the Spin training in Appendix~\ref{subsec:app-impl-spin}, 8.2k hours of data are used during DS.

\subsection{Low-budget Robust data2vec}
\label{subsec:app-impl-rd2v}

We follow the implementation of \citet{zhu2023robust} with fairseq~\citep{ott2019fairseq}.\footnote{https://github.com/zqs01/data2vecnoisy}
We changed the training data from CHiME-4 to LibriSpeech for a fair comparison with our method.
Because we found a long training schedule is necessary for Robust data2vec converge, the number of updates is the same as the original implementation~(100k).
Meanwhile, the mini-batch size is reduced from 63 to 6.25 minutes so that the amount of speech data processed is similar to \proposed.
The rest of the hyperparameters remain the same since we found the original ones are sufficiently good.
As shown in Table~\ref{tab:pr-asr}, the low-budget Robust data2vec model has a significant performance degradation compared with the fully-trained version, implying the necessity to train this model with a large batch size.
Concurrently, \proposed~achieves superior results under the same budget, indicating that our approach is more data-efficient.

We found that the low-budget Robust data2vec is particularly difficult to train.
First, \citet{hsu2021hubert} has shown that large batch sizes favor speech SSL model training, which applies to Robust data2vec.
Second, we found that the low-budget model fails when trained with CHiME-4, indicating the training corpus highly affects model convergence.
Third, we did not have the resources to find the optimal hyperparameters.
However, the hyperparameters for data2vec must be carefully determined to let the model converge~\citep{baevski2022data2vec}, which is amplified when the training scale is reduced.
In conclusion, we were unable to find a setup where a comparable computational budget is used for both R-Spin and Robust data2vec.
Nonetheless, the results demonstrate that the proposed R-Spin is much easier to operate under low-budget scenarios.

\subsection{Phoneme Recognition}
\label{subsec:app-impl-pr}

We follow the setup in SUPERB~\citep{yang2021superb}, which freezes each SSL model and uses a set of learnable weights to weighted-sum hidden features of all layers.
The aggregated frame-wise features are fed into a lightweight linear prediction head to perform downstream tasks.
Only the prediction head and the weighted-sum mechanism are fine-tuned with clean and labeled speech data to reveal the capabilities of SSL models.
The LibriSpeech train-clean-100 and the test-other subsets are used as the training and evaluation datasets, respectively.
The prediction head projects features to phoneme labels.
Unlike the SUPERB setup, the learning rate is 5 $\times$ 10\textsuperscript{$-$4}~(originally 10\textsuperscript{$-$2}) to obtain a better performance, and the number of training updates is 30k~(originally 100k).
The noise and perturbation data sources are listed as follows.
\begin{enumerate}[nolistsep]
    \item \textbf{Gaussian Noise}:
        The Gaussian noises are generated with PyTorch.
    \item \textbf{Background Noise}:
        The background noises are sampled from the MUSAN dataset.
        We duplicate the noise recording when it is shorter than the input.
        Otherwise, we randomly crop the recording to match the utterance length.
    \item \textbf{Reverberation}:
        We filter waveforms with real and simulated impulse responses in RIRS~\citep{ko2017rirs}.\footnote{Released under Apache 2.0}
        The scores for the real and simulated reverberation are averaged.
\end{enumerate}

\subsection{CHiME-4 ASR}
\label{subsec:app-impl-asr}

We follow the ASR task of SUPERB, but the prediction heads~(two-layer BLSTM) are trained with the clean portion of the CHiME-4 speech corpus obtained from the WSJ0 corpus~\citep{Paul92wsj}, consisting of 14 hours of clean English speech.
The number of training updates is 100k~(originally 200k).
The trained ASR models are evaluated on the 1-channel track of the CHiME-4 challenge.
We report the averaged WERs over each subset~(real and simulated data).
We apply Whisper normalization to all ASR results for a fair comparison with the Whisper toplines.\footnote{https://github.com/openai/whisper}

\subsection{Acoustic Pieces}
\label{subsec:app-ap}
\input{fig/ap_vocab_size}

We implemented the BPE algorithm in Python.
The AP vocabulary sizes vs. the actual APs used are shown in Fig.~\ref{fig:ap-vocab}.
Since some merging operations in BPE replace previously learned BPE vocabularies with new ones, the number of used BPEs in the encoded LibriSpeech corpus is smaller than the learned BPE vocabularies.
E.g., we have a sentence ``\texttt{a a b b a}'' and the learned BPE vocabularies \texttt{a}, \texttt{b}, \texttt{aa}, \texttt{bb}, and \texttt{aabb}.
Then, the encoded sentence is ``\texttt{aabb a},'' eliminating the \textit{intermediate} vocabularies \texttt{b}, \texttt{aa}, and \texttt{bb}.
Thus, the number of classes in the linear prediction head for $\mathcal{L}_{\text{Aux}}$ is adjusted accordingly.
E.g., the prediction head's output for R-Spin\textsubscript{32, 40k} is 19857 instead of 40000.

\section{Additional Experiments}
\label{sec:app-exp}

\input{table/ablation_appendix}

\subsection{Ablation Studies}
\label{subsec:app-ablation}

\subsubsection{Hyperparameters}
To examine the impact of the auxiliary loss, we change the value of $\lambda$ in Eq.~\ref{eq:total-loss}.
As shown in Table~\ref{tab:ablation-appendix}, the differences of ASR WERs between different $\lambda$'s are negligible.
We can conclude that combining $\mathcal{L}_{\text{Spin}}$ and $\mathcal{L}_{\text{Aux}}$ is necessary, and the ratio between the two objectives is robust.

\subsubsection{Layer to Apply $\mathcal{L}_{\text{Aux}}$}
In the R-Spin design, $\mathcal{L}_{\text{Aux}}$ is applied to the last layer.
Here, we apply $\mathcal{L}_{\text{Aux}}$ to other hidden layers to verify that our approach leads to the best overall result.
When we move the auxiliary loss $\mathcal{L}_{\text{Aux}}$ to lower layers, the performance degrades significantly, showing that this loss should regularize the entire model.
Otherwise, the Spin loss still makes the codebook collapse.

\subsubsection{Layer to Apply $\mathcal{L}_{\text{Spin}}$}
Similar to the previous experiments, we apply $\mathcal{L}_{\text{Spin}}$ to lower layers to find the optimal design.
The ASR performance degrades slightly when we move the Spin objective function to lower layers.
With the results of $\mathcal{L}_{\text{Aux}}$, we conclude that a relatively good strategy for applying the two proposed loss functions is adding both to the top layer.

\subsubsection{Fine-tuned Layers}
Here, we inspect the benefits of fine-tuning SSL models entirely in contrast to Spin, which fine-tunes only the top two layers.
Hence, we reduce the number of fine-tuned layers.
The results indicate that the model cannot adapt to noisy scenarios by fine-tuning fewer top layers.
Thus, R-Spin is beneficial since we can now fine-tune the entire model for noisy conditions.

\subsubsection{Data}
We further changed the data for R-Spin DS to reveal the impact of training corpora on performance.
We found that WERs degrade slightly when the training corpus size is reduced.
Moreover, the ASR performance degrades prominently by increasing the SNRs of the background noise for the noise-invariant training.
Hence, the choice of noise data and SNRs has a greater impact on the downstream performance than that of the clean speech corpus.

\input{fig/superb_weight}
\input{table/phone_seg}

\subsection{Importance of Hidden Representations}
\label{subsec:app-hid-repr}

We visualize the weighted sum mechanism for phoneme and speech recognition to understand the importance of each layer.
The weights form a probability distribution over all layers~(including the CNN feature extractor).
The features of each layer are weighted and summed with these weights.
However, the scale of the embedding spaces differs between layers.
Suppose the weight of a layer is small, but the norm of the corresponding hidden vectors is large.
That layer might contribute significantly to the downstream task.
Consequently, we normalize each weight by multiplying with the averaged L2 norm of the corresponding layer embedding, which is written as
\begin{equation*}
    \hat{w}_l = w_l \cdot \mathbb{E} \left[ \left\| \boldsymbol{h}^{(l)} \right\|_2 \right],
\end{equation*}
where $w_l$ and $\boldsymbol{h}^{(l)}$ are respectively the unnormalized weight and hidden features of layer $l$, and $\mathbb{E}$ is the expectation over all samples from the LibriSpeech dev-clean and dev-other sets~\citep{chang2022distilhubert}.
Next, the new set of weights $\hat{w}_l$ is normalized to sum to one.
As shown in Fig.~\ref{fig:superb-weight}, the last layer of R-Spin has the least speaker and noise information, but the second last layer offers the best phoneme representations.
In contrast, when R-Spin is applied, the best ASR layers tend to shift towards the last layer.

\subsection{Unsupervised Phoneme Segmentation}
\label{subsec:app-segment}

This section inspects the phoneme segmentation capability of the proposed methods.
As shown in Table~\ref{tab:phone-seg}, segmenting speech with Spin codebook or acoustic pieces is inferior to prior methods specifically designed for phoneme segmentation because no explicit constraints are added to encourage phoneme boundary detection.
Still, some \proposed~discrete units like \proposed\textsubscript{32, AP40k} surpass the oracle uniform baseline, indicating that the discrete unit boundaries are close to phoneme boundaries.
The results align with the findings in Fig.~\ref{fig:bpe-phone}.

\clearpage

\input{table/ssl_costs}
\input{table/computation}

\section{Computation Resources}
\label{sec:app-compute-resource}

The costs of self-supervised pre-training and domain-specific self-supervision methods are shown in Table~\ref{tab:ssl-costs}.
The required computation resources for each training task in this paper are listed in Table~\ref{tab:computation}.
Note that all results in this paper are obtained with a single run.
\footnotetext{The model checkpoints will be made public on https://github.com/vectominist/spin}

\section{t-SNE of Hidden Representations}
\label{sec:app-tsne-repr}

We plot more t-SNE visualization of hidden representations in Figs.~\ref{fig:trajectory-cross-dialect}, \ref{fig:trajectory-mspkr}, \ref{fig:trajectory-hubert}, \ref{fig:trajectory-rspin}, \ref{fig:trajectory-hubert-domain}, and \ref{fig:trajectory-rspin-domain}.

\input{fig/trajectory_dialect}
\input{fig/trajectory_mspkr}

\input{fig/trajectory_hubert_full}
\input{fig/trajectory_rspin_full}

\input{fig/trajectory_hubert_domain_full}
\input{fig/trajectory_rspin_domain_full}

%% file: fig/ap_vocab_size.tex
\begin{figure}[t]
    \centering
    \includegraphics[width=0.9\linewidth]{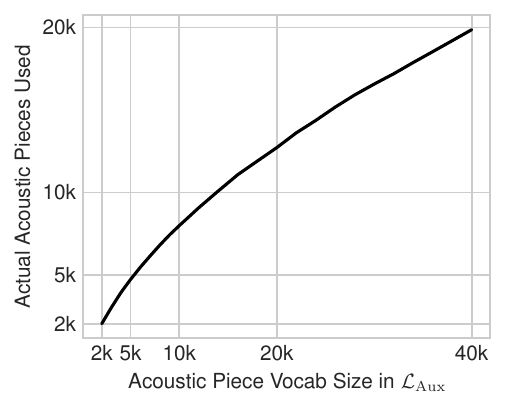}
    \vspace{-5pt}
    \caption{AP size vs. actual vocabularies used.}
    \label{fig:ap-vocab}
\end{figure}

%% file: table/ablation_appendix.tex
\begin{table}[t]
    \centering
    \begin{adjustbox}{max width=0.9\linewidth}
    \begin{tabular}{l@{~~~~~~~~}cc}
        \toprule
        & \multicolumn{2}{c}{CHiME-4} \\
        Method & Real & Sim \\
        \midrule
        Spin\textsubscript{2048}~\citep{chang2023spin} & 52.1 & 46.6 \\
        \proposed\textsubscript{32, BPE40k} & 26.4 & 26.6 \\
        \midrule
        \multicolumn{3}{l}{\textbf{Hyperparameters}} \\
        ~~$\lambda =$ 1 & 26.3 & 27.7 \\
        ~~$\lambda =$ 0.5 & 26.6 & 27.3 \\
        \midrule
        \multicolumn{3}{l}{\textbf{Layer to Apply} $\mathcal{L}_{\text{Aux}}$} \\
        ~~Layer 11 & 28.1 & 28.8 \\
        ~~Layer 10 & 34.7 & 33.8 \\
        \midrule
        \multicolumn{3}{l}{\textbf{Layer to Apply} $\mathcal{L}_{\text{Spin}}$} \\
        ~~Layer 11 & 27.2 & 27.9 \\
        ~~Layer 10 & 27.0 & 27.8 \\
        \midrule
        \multicolumn{3}{l}{\textbf{Fine-tuned Layers}} \\
        ~~Top 10 Layers & 29.7 & 30.0 \\
        ~~Top 6 Layers & 39.4 & 37.5 \\
        \midrule
        \multicolumn{3}{l}{\textbf{Dataset}} \\
        ~~LibriSpeech 100h & 27.2 & 27.6 \\
        ~~LibriSpeech 360h & 26.6 & 27.6 \\
        \midrule
        \multicolumn{3}{l}{\textbf{Noise SNR Range}} \\
        ~~0 -- 20dB & 29.0 & 28.6 \\
        \bottomrule
    \end{tabular}
    \end{adjustbox}
    \caption{CHiME-4 ASR results for additional ablation studies based on fine-tuned WavLM models.}
    \label{tab:ablation-appendix}
\end{table}

%% file: fig/superb_weight.tex
\begin{figure*}[t]
    \centering
    \begin{subfigure}[b]{0.495\linewidth}
        \centering
        \includegraphics[width=\linewidth]{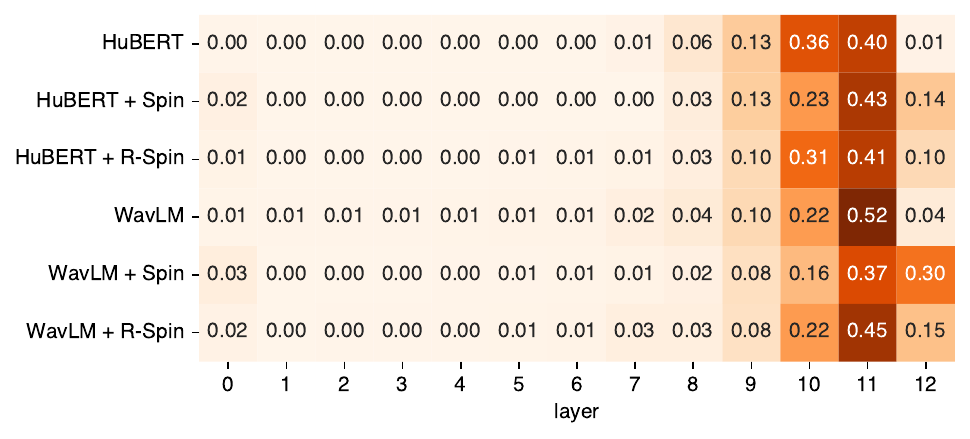}
        \vspace{-20pt}
        \caption{Phoneme Recognition}
        \label{fig:weight-pr}
    \end{subfigure}
    \hfill
    \begin{subfigure}[b]{0.495\linewidth}
        \centering
        \includegraphics[width=\linewidth]{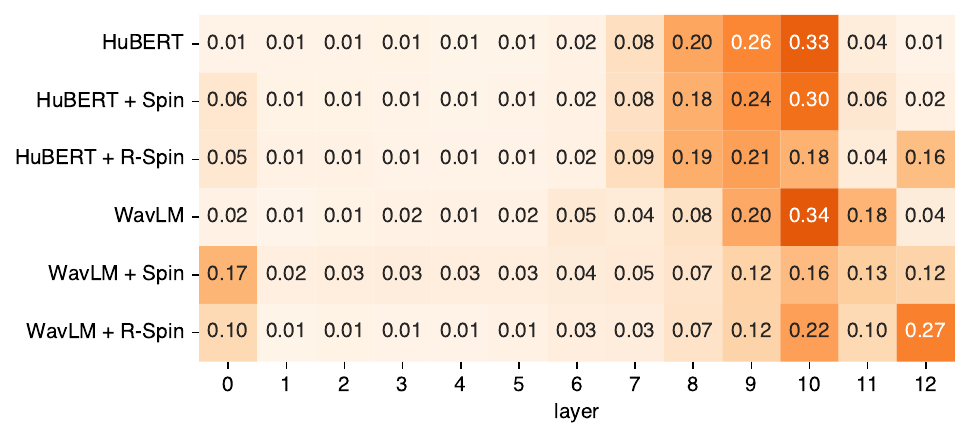}
        \vspace{-20pt}
        \caption{Automatic Speech Recognition}
        \label{fig:weight-asr}
    \end{subfigure}
    \caption{
        Normalized weights of the weighted sum mechanism in the SUPERB PR and ASR.
    }
    \label{fig:superb-weight}
\end{figure*}

%% file: table/phone_seg.tex
\begin{table*}[t!]
    \centering
    \begin{adjustbox}{max width=0.9\linewidth}
    \begin{threeparttable}
    \begin{tabular}{lccccc}
        \toprule
        Method & Precision$\uparrow$ & Recall$\uparrow$ & F1$\uparrow$ & OS$\rightarrow$0 & R-val$\uparrow$ \\
        \midrule
        \textbf{Baseline} \\
        ~~Oracle Uniform & 56.49 & 62.99 & 59.56 & 11.50 & 63.47 \\
        \midrule
        \textbf{Unsupervised} \\
        ~~CPC~\citep{kreuk2020self} & 83.89 & 83.55 & 83.71 &  & 86.02 \\
        ~~SCPC~\citep{bhati2021segmental} & 84.63 & 86.04 & 85.33 &  & 87.44 \\
        ~~HuBERT readout~\citep{strgar2023phoneme} & \textbf{90.98} & \textbf{88.48} & \textbf{89.71} &  & \textbf{90.98} \\
        \midrule
        \textbf{Spin Codebook} \\
        ~~HuBERT + Spin\textsubscript{128}~\citep{chang2023spin} & 64.76 & 87.87 & 74.56 & 35.69 & 64.25 \\
        ~~HuBERT + Spin\textsubscript{256}~\citep{chang2023spin} & 61.71 & 90.84 & 73.49 & 47.22 & 56.02 \\
        ~~HuBERT + Spin\textsubscript{512}~\citep{chang2023spin} & 60.78 & 95.46 & 74.27 & 57.07 & 49.60 \\
        ~~HuBERT + Spin\textsubscript{1024}~\citep{chang2023spin} & 59.93 & 97.95 & 74.36 & 63.44 & 45.11 \\
        ~~HuBERT + Spin\textsubscript{2048}~\citep{chang2023spin} & 58.58 & \textbf{99.46} & 73.73 & 69.77 & 40.26 \\
        ~~HuBERT + Spin\textsubscript{2048} (for AP) & 61.31 & 96.87 & \textbf{75.09} & 58.00 & 49.34 \\
        ~~HuBERT + R-Spin\textsubscript{32, AP40k} & 64.73 & 71.47 & 67.93 & \textbf{10.41} & 71.05 \\
        ~~WavLM + R-Spin\textsubscript{16, AP40k} & 60.73 & 68.22 & 64.26 & 12.32 & 67.36 \\
        ~~WavLM + R-Spin\textsubscript{32, AP40k} & \textbf{65.12} & 73.76 & 69.17 & 13.28 & \textbf{71.33} \\
        ~~WavLM + R-Spin\textsubscript{64, AP40k} & 63.02 & 73.63 & 67.91 & 16.83 & 69.08 \\
        ~~WavLM + R-Spin\textsubscript{128, AP40k} & 61.44 & 72.42 & 66.48 & 17.88 & 67.49 \\
        ~~WavLM + R-Spin\textsubscript{256, AP40k} & 60.80 & 78.61 & 68.57 & 29.28 & 63.95 \\
        ~~WavLM + R-Spin\textsubscript{512, AP40k} & 59.66 & 82.94 & 69.40 & 39.01 & 58.89 \\
        ~~WavLM + R-Spin\textsubscript{1024, AP40k} & 59.03 & 89.09 & 71.01 & 50.94 & 52.09 \\
        ~~WavLM + R-Spin\textsubscript{2048, AP40k} & 58.47 & 94.19 & 72.15 & 61.08 & 45.67 \\
        \midrule
        \textbf{Acoustic Pieces} \\
        ~~HuBERT + Spin\textsubscript{2048} AP5k & 60.80 & \textbf{71.64} & \textbf{65.77} & 17.83 & 66.92 \\
        ~~HuBERT + Spin\textsubscript{2048} AP10k & 61.37 & 68.51 & 64.74 & 11.64 & 67.97 \\
        ~~HuBERT + Spin\textsubscript{2048} AP20k & 61.76 & 68.74 & 65.06 & 11.29 & 68.34 \\
        ~~HuBERT + Spin\textsubscript{2048} AP40k & \textbf{62.10} & 68.54 & 65.16 & \textbf{10.37} & \textbf{68.65} \\
        \bottomrule
    \end{tabular}
    \end{threeparttable}
    \end{adjustbox}
    \caption{
        Unsupervised phoneme segmentation on TIMIT test set.
        OS and R-val respectively denote the over-segmentation rate and R-value~\citep{rasanen2009rval}.
        Oracle uniform is a segmentation method that splits speech into equal-length segments, given the ground truth number of phoneme boundaries.
        Unknown results are left blank.
    }
    \label{tab:phone-seg}
\end{table*}

%% file: table/ssl_costs.tex
\begin{table*}[t]
    \centering
    \begin{adjustbox}{max width=0.93\linewidth}
    \begin{tabular}{@{~}l@{~~}c@{~~}c@{~~}c@{~~}c@{~~}c@{~~}c@{~~}c@{~}}
        \toprule
        & & & \multirow{3}{*}{\shortstack[c]{~\\~\\Batch \\Size\\ (minutes)}} & \multirow{3}{*}{\shortstack[c]{~\\Processed\\ Speech \\(hours)}} & & &  \\
        & & & & & & \multirow{2}{*}{\shortstack[c]{~\\~\\GPU\\ Hours}} & \multirow{2}{*}{\shortstack[c]{~\\Open\\ Model}} \\
        Model & Init & Updates & & & \#GPUs &  \\
        \midrule
        \textbf{Self-supervised Pre-training (Clean Speech)} \\
        ~~wav2vec 2.0~\cite{baevski2020wav2vec2} & -- & 400k & 96 & 640k & 64 & 2458 & \Checkmark \\
        ~~HuBERT~\citep{hsu2021hubert} & -- & 250k + 400k & 47 & 505k & 32 & 1976 & \Checkmark \\
        ~~data2vec~\citep{baevski2022data2vec} & -- & 400k & 63 & 420k & 16 &  & \Checkmark \\
        ~~DinoSR~\citep{liu2023dinosr} & -- & 400k & 63 & 420k & 16 & 2880 & \Checkmark \\
        \midrule
        \textbf{Self-supervised Pre-training (Noisy Speech)} \\
        ~~WavLM~\citep{chen2022wavlm} & -- & 250k + 400k & 187 & 1439k & 32 &  & \Checkmark \\
        ~~wav2vec-Switch~\citep{wang2022wav2vec-switch} & -- & 400k & 96 & 640k & 32 &  & \XSolidBrush \\
        ~~SPIRAL~\citep{huang2022spiral} & -- & 200k & 100 & 333k & 16 & 499 & \Checkmark \\
        \midrule
        \textbf{Domain-specific Self-supervision} \\
        ~~ContentVec~\citep{qian2022contentvec} & HuBERT & 100k & 46 & 76k & 36 & 684 & \Checkmark \\
        ~~HuBERT-MGR~\citep{huang2022improving} & HuBERT & 400k & 12 & 78k & 8 & 768 & \Checkmark \\
        ~~Robust data2vec~\citep{zhu2023robust} & data2vec & 100k & 63 & 105k & 16 &  & \Checkmark \\
        ~~deHuBERT~\citep{ng2023dehubert} & HuBERT & 250k &  &  &  &  & \XSolidBrush \\
        ~~Spin\textsubscript{2048}~\citep{chang2023spin} & HuBERT & 5k & 43 & 0.4k & 1 & 1 & \Checkmark \\
        \midrule
        \textbf{This Paper} \\
        ~~Robust data2vec (low budget) & data2vec & 100k & 6.3 & 10.4k & 2 & 44 & $\triangle$ \\
        ~~Spin\textsubscript{2048} (for AP40k) & HuBERT & 10k & 43 & 7.1k & 2 & 8 & $\triangle$ \\
        ~~R-Spin\textsubscript{32, AP40k} & HuBERT & 10k & 6.4 & 1.1k & 2 & 16 & $\triangle$ \\
        \bottomrule
    \end{tabular}
    \end{adjustbox}
    \vspace{-3pt}
    \caption{
        SSL and DS costs of models with 95M parameters.
        The ``Init'' column shows the pre-trained models used for initialization.
        $\triangle$ denotes models in this paper, which will be made publicly available in the near future.\footnotemark~
        Note that duplicate input utterances by data augmentation are not included when calculating the hours of speech processed.
        The number of GPU hours required for training is roughly estimated, so the true values might differ slightly.
        The availability of the models listed was updated in March 2024.
        Unknown data are left blank.
    }
    \label{tab:ssl-costs}
\end{table*}

%% file: table/computation.tex
\begin{table}[t]
    \centering
    \begin{adjustbox}{max width=\linewidth}
    \begin{tabular}{l@{~~~}c@{~~~}c@{~~~}c}
        \toprule
        Task & Updates & Hours & GPU \\
        \midrule
        (\ref{subsec:app-impl-spin}) Spin & 10k & 4 & A6000$\times$2 \\
        (\ref{subsec:app-impl-rspin}) R-Spin & 10k & 8 & A6000$\times$2 \\
        (\ref{subsec:app-impl-rd2v}) Robust data2vec & 100k & 22 & A6000$\times$2 \\
        (\ref{subsec:app-impl-pr}) SUPERB PR & 30k & 10 & 2080 Ti \\
        (\ref{subsec:app-impl-asr}) SUPERB ASR & 100k & 20 & A5000 \\
        (\ref{subsubsec:exp-spk-inv}) SUPERB SID & 50k & 4 & A6000 \\
        \bottomrule
    \end{tabular}
    \end{adjustbox}
    \caption{Computation resources used in the experiments.}
    \label{tab:computation}
\end{table}

%% file: fig/trajectory_dialect.tex
\begin{figure*}
    \centering
    \begin{subfigure}[b]{0.45\linewidth}
        \centering
        \captionsetup{justification=centering}
        \includegraphics[width=\linewidth]{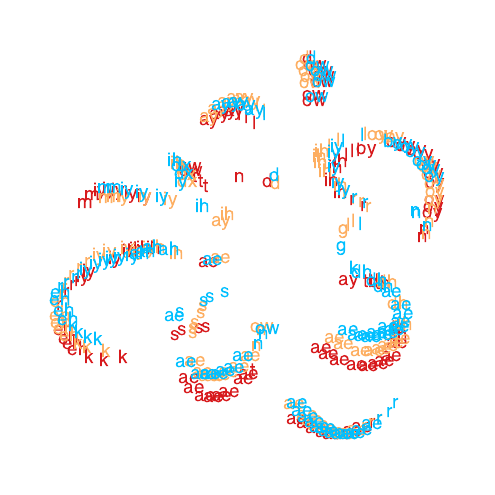}
        \vspace{-18pt}
        \caption{HuBERT\\Layer 9}
        \label{fig:traj-hubert-dialect}
    \end{subfigure}
    \begin{subfigure}[b]{0.45\linewidth}
        \centering
        \captionsetup{justification=centering}
        \includegraphics[width=\linewidth]{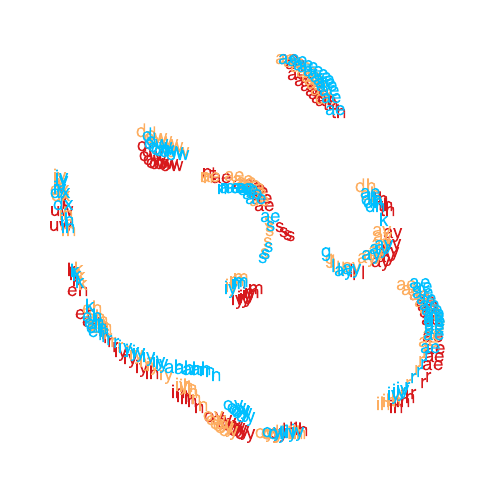}
        \vspace{-18pt}
        \caption{HuBERT + R-Spin\\Layer 12}
        \label{fig:traj-rspin-dialect}
    \end{subfigure}
    \caption{
        t-SNE visualization of three speakers with different English dialects~(see Fig.~\ref{fig:trajectory} for details).
    }
    \label{fig:trajectory-cross-dialect}
\end{figure*}


%% file: fig/trajectory_mspkr.tex
\begin{figure*}
    \centering
    \begin{subfigure}[b]{0.45\linewidth}
        \centering
        \captionsetup{justification=centering}
        \includegraphics[width=\linewidth]{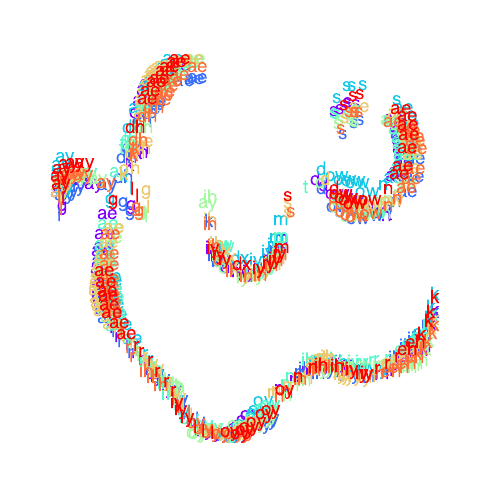}
        \vspace{-18pt}
        \caption{HuBERT\\Layer 9}
        \label{fig:traj-hubert-mspkr}
    \end{subfigure}
    \begin{subfigure}[b]{0.45\linewidth}
        \centering
        \captionsetup{justification=centering}
        \includegraphics[width=\linewidth]{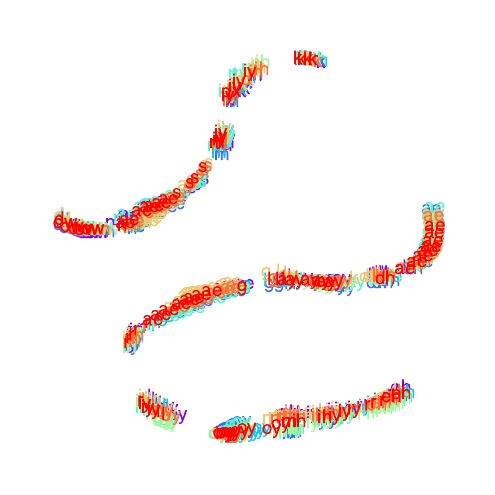}
        \vspace{-18pt}
        \caption{HuBERT + R-Spin\\Layer 12}
        \label{fig:traj-rspin-mspkr}
    \end{subfigure}
    \caption{
        t-SNE visualization of eight speakers with different English dialects~(see Fig.~\ref{fig:trajectory} for details).
    }
    \label{fig:trajectory-mspkr}
\end{figure*}


%% file: fig/trajectory_hubert_full.tex
\begin{figure*}
    \centering
    \begin{subfigure}[b]{0.325\linewidth}
        \centering
        \includegraphics[width=\linewidth]{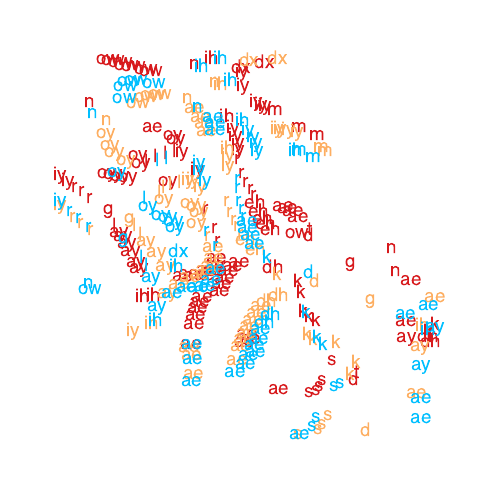}
        \vspace{-22pt}
        \caption{Layer 1}
    \end{subfigure}
    \begin{subfigure}[b]{0.325\linewidth}
        \centering
        \includegraphics[width=\linewidth]{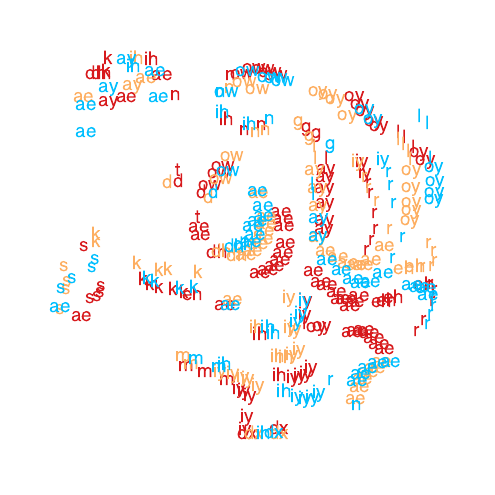}
        \vspace{-22pt}
        \caption{Layer 2}
    \end{subfigure}
    \begin{subfigure}[b]{0.325\linewidth}
        \centering
        \includegraphics[width=\linewidth]{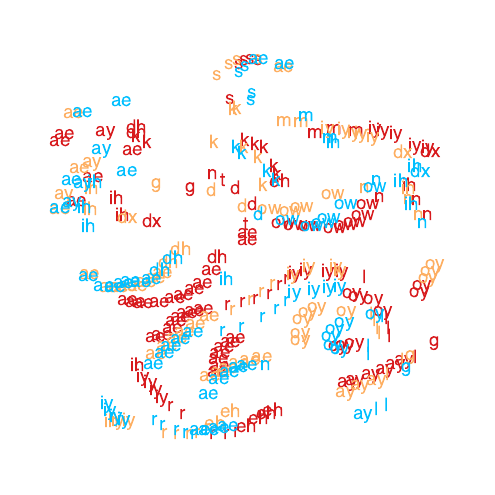}
        \vspace{-22pt}
        \caption{Layer 3}
    \end{subfigure}
    \begin{subfigure}[b]{0.325\linewidth}
        \centering
        \includegraphics[width=\linewidth]{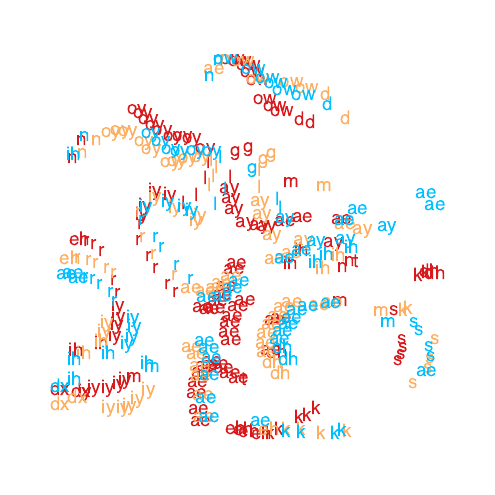}
        \vspace{-22pt}
        \caption{Layer 4}
    \end{subfigure}
    \begin{subfigure}[b]{0.325\linewidth}
        \centering
        \includegraphics[width=\linewidth]{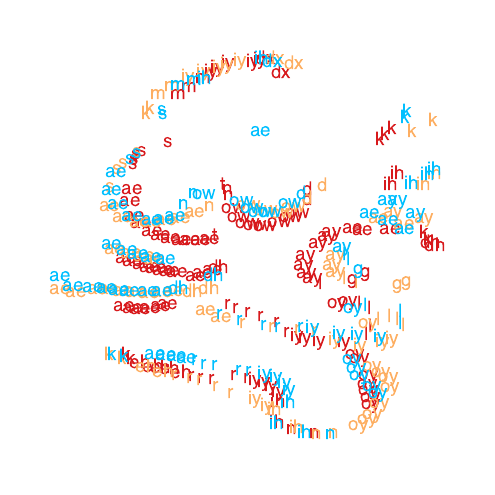}
        \vspace{-22pt}
        \caption{Layer 5}
    \end{subfigure}
    \begin{subfigure}[b]{0.325\linewidth}
        \centering
        \includegraphics[width=\linewidth]{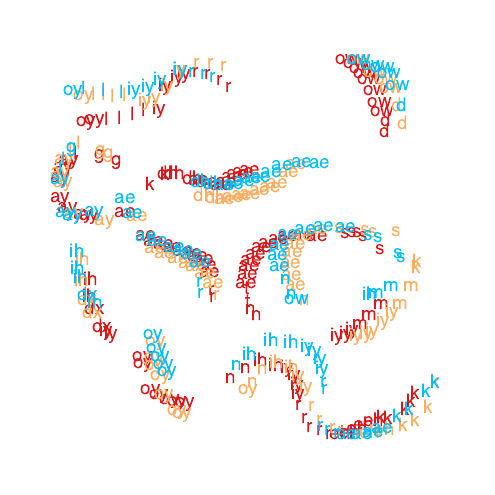}
        \vspace{-22pt}
        \caption{Layer 6}
    \end{subfigure}
    \begin{subfigure}[b]{0.325\linewidth}
        \centering
        \includegraphics[width=\linewidth]{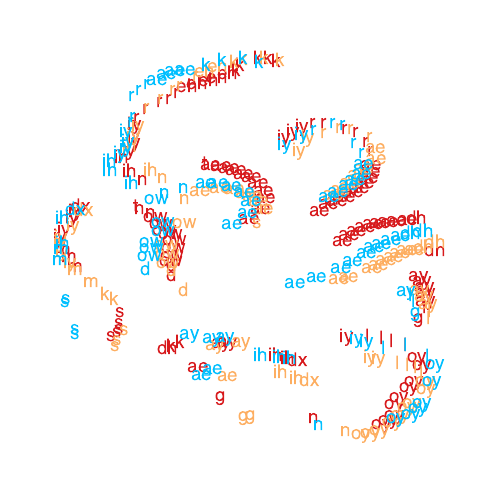}
        \vspace{-22pt}
        \caption{Layer 7}
    \end{subfigure}
    \begin{subfigure}[b]{0.325\linewidth}
        \centering
        \includegraphics[width=\linewidth]{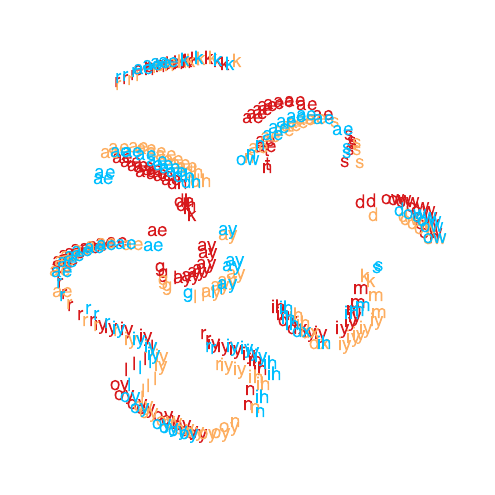}
        \vspace{-22pt}
        \caption{Layer 8}
    \end{subfigure}
    \begin{subfigure}[b]{0.325\linewidth}
        \centering
        \includegraphics[width=\linewidth]{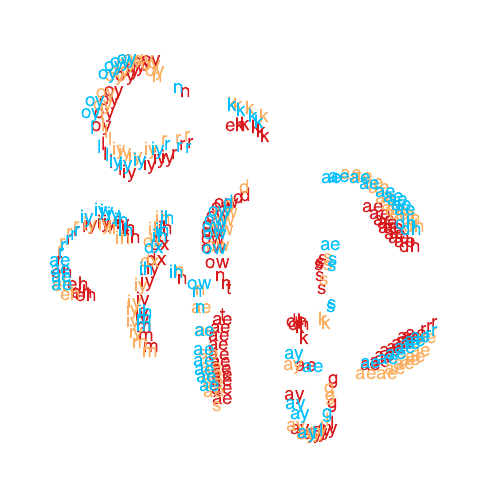}
        \vspace{-22pt}
        \caption{Layer 9}
    \end{subfigure}
    \begin{subfigure}[b]{0.325\linewidth}
        \centering
        \includegraphics[width=\linewidth]{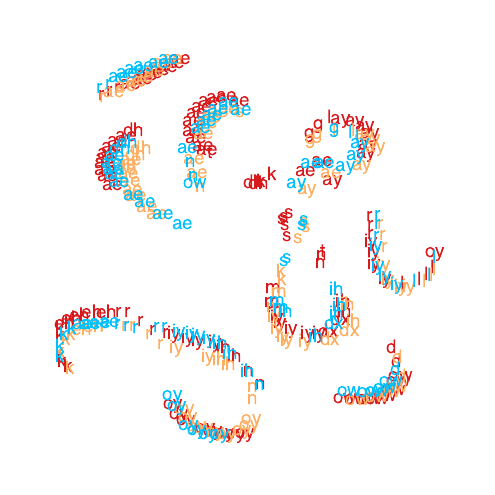}
        \vspace{-22pt}
        \caption{Layer 10}
    \end{subfigure}
    \begin{subfigure}[b]{0.325\linewidth}
        \centering
        \includegraphics[width=\linewidth]{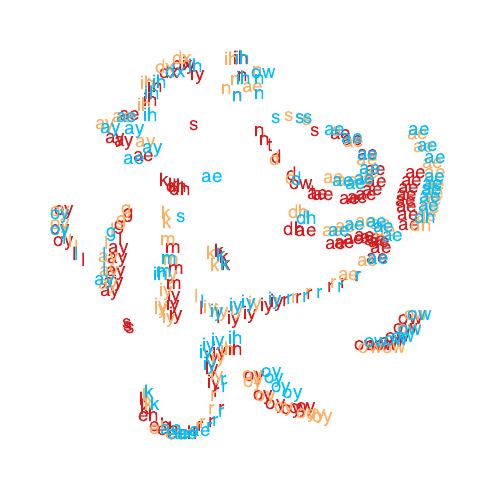}
        \vspace{-22pt}
        \caption{Layer 11}
    \end{subfigure}
    \begin{subfigure}[b]{0.325\linewidth}
        \centering
        \includegraphics[width=\linewidth]{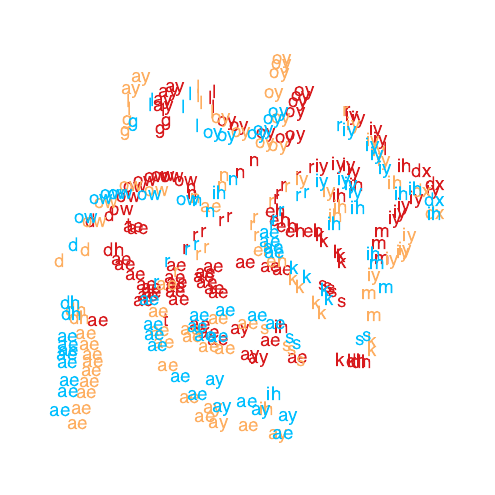}
        \vspace{-22pt}
        \caption{Layer 12}
    \end{subfigure}
    \caption{
        t-SNE visualization of HuBERT representations of the same utterance spoken by three speakers~(see Fig.~\ref{fig:trajectory} for details).
    }
    \label{fig:trajectory-hubert}
    \vspace{-6pt}
\end{figure*}

%% file: fig/trajectory_rspin_full.tex
\begin{figure*}
    \centering
    \begin{subfigure}[b]{0.325\linewidth}
        \centering
        \includegraphics[width=\linewidth]{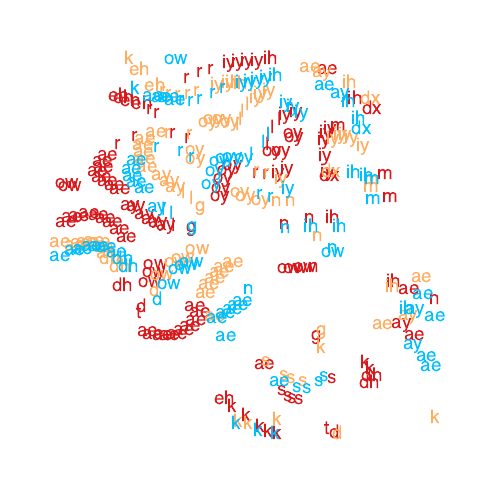}
        \vspace{-22pt}
        \caption{Layer 1}
    \end{subfigure}
    \begin{subfigure}[b]{0.325\linewidth}
        \centering
        \includegraphics[width=\linewidth]{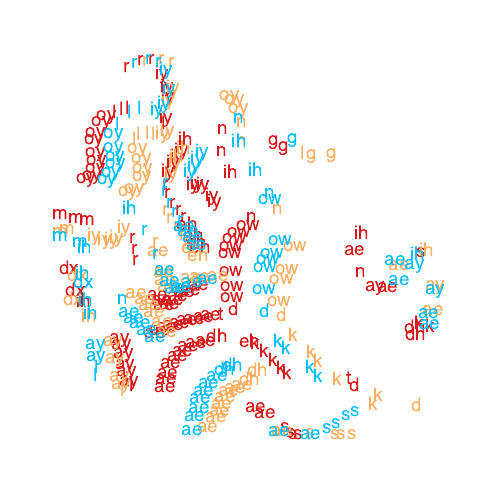}
        \vspace{-22pt}
        \caption{Layer 2}
    \end{subfigure}
    \begin{subfigure}[b]{0.325\linewidth}
        \centering
        \includegraphics[width=\linewidth]{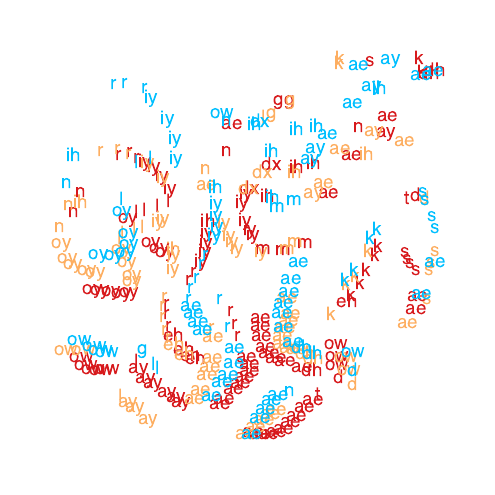}
        \vspace{-22pt}
        \caption{Layer 3}
    \end{subfigure}
    \begin{subfigure}[b]{0.325\linewidth}
        \centering
        \includegraphics[width=\linewidth]{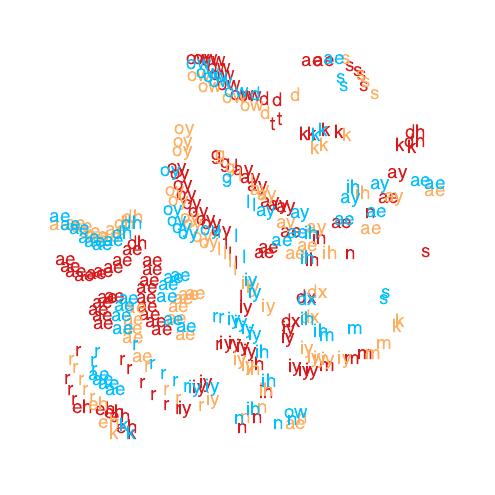}
        \vspace{-22pt}
        \caption{Layer 4}
    \end{subfigure}
    \begin{subfigure}[b]{0.325\linewidth}
        \centering
        \includegraphics[width=\linewidth]{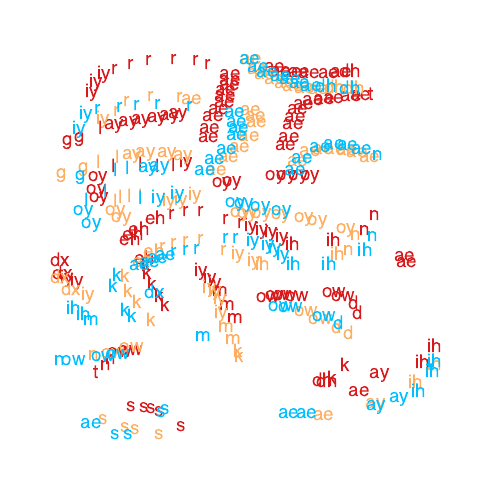}
        \vspace{-22pt}
        \caption{Layer 5}
    \end{subfigure}
    \begin{subfigure}[b]{0.325\linewidth}
        \centering
        \includegraphics[width=\linewidth]{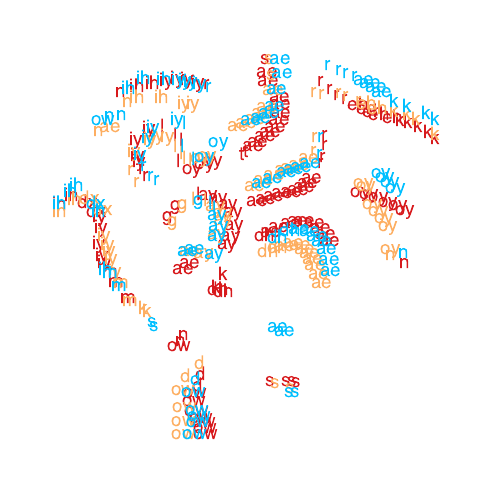}
        \vspace{-22pt}
        \caption{Layer 6}
    \end{subfigure}
    \begin{subfigure}[b]{0.325\linewidth}
        \centering
        \includegraphics[width=\linewidth]{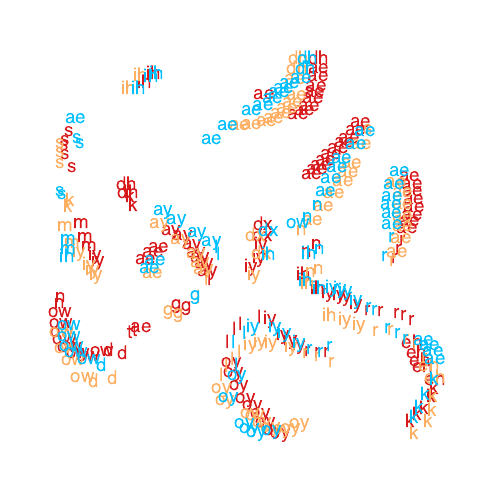}
        \vspace{-22pt}
        \caption{Layer 7}
    \end{subfigure}
    \begin{subfigure}[b]{0.325\linewidth}
        \centering
        \includegraphics[width=\linewidth]{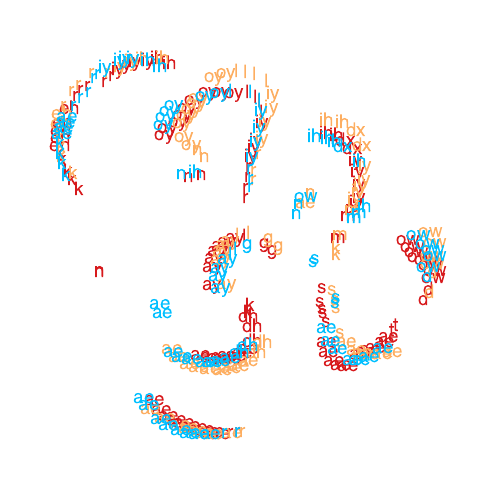}
        \vspace{-22pt}
        \caption{Layer 8}
    \end{subfigure}
    \begin{subfigure}[b]{0.325\linewidth}
        \centering
        \includegraphics[width=\linewidth]{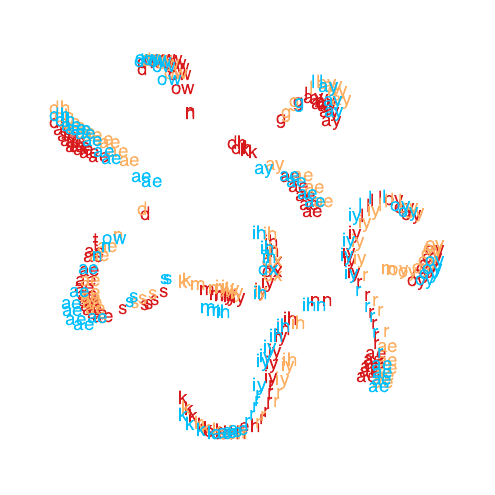}
        \vspace{-22pt}
        \caption{Layer 9}
    \end{subfigure}
    \begin{subfigure}[b]{0.325\linewidth}
        \centering
        \includegraphics[width=\linewidth]{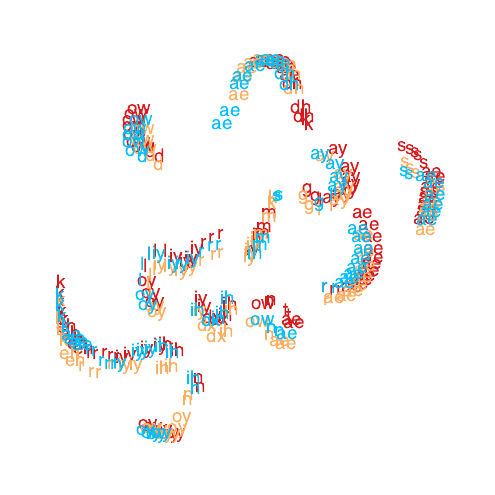}
        \vspace{-22pt}
        \caption{Layer 10}
    \end{subfigure}
    \begin{subfigure}[b]{0.325\linewidth}
        \centering
        \includegraphics[width=\linewidth]{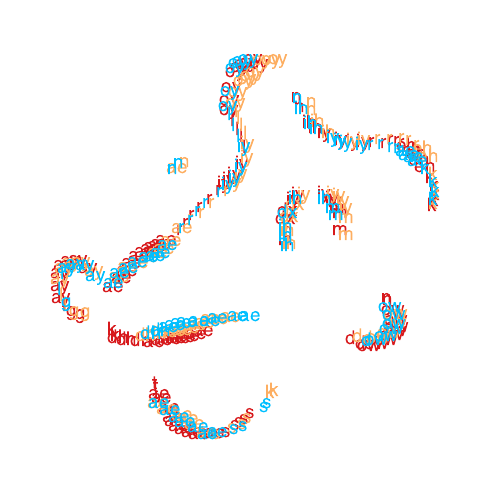}
        \vspace{-22pt}
        \caption{Layer 11}
    \end{subfigure}
    \begin{subfigure}[b]{0.325\linewidth}
        \centering
        \includegraphics[width=\linewidth]{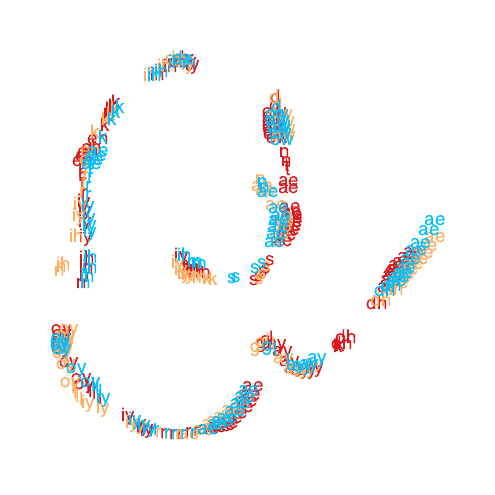}
        \vspace{-22pt}
        \caption{Layer 12}
    \end{subfigure}
    \caption{
        t-SNE visualization of HuBERT + \proposed~representations of the same utterance spoken by three speakers~(see Fig.~\ref{fig:trajectory} for details).
    }
    \label{fig:trajectory-rspin}
    \vspace{-6pt}
\end{figure*}

%% file: fig/trajectory_hubert_domain_full.tex
\begin{figure*}
    \centering
    \begin{subfigure}[b]{0.300\linewidth}
        \centering
        \includegraphics[width=\linewidth]{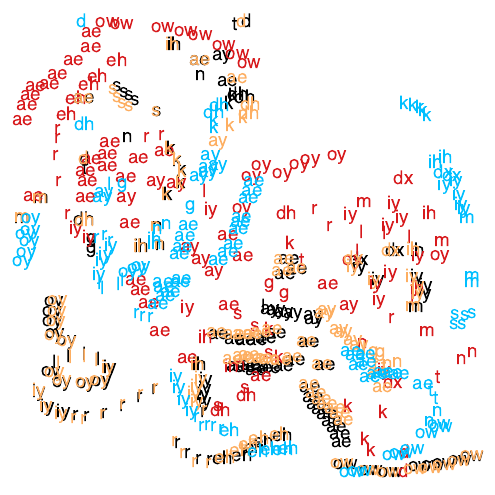}
        \vspace{-18pt}
        \caption{Layer 1}
    \end{subfigure}
    \hfill
    \begin{subfigure}[b]{0.300\linewidth}
        \centering
        \includegraphics[width=\linewidth]{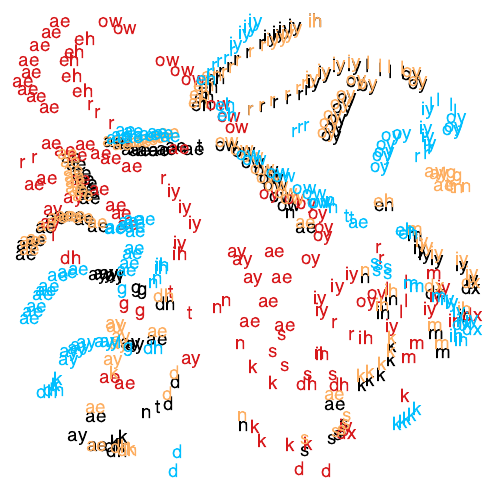}
        \vspace{-18pt}
        \caption{Layer 2}
    \end{subfigure}
    \hfill
    \begin{subfigure}[b]{0.300\linewidth}
        \centering
        \includegraphics[width=\linewidth]{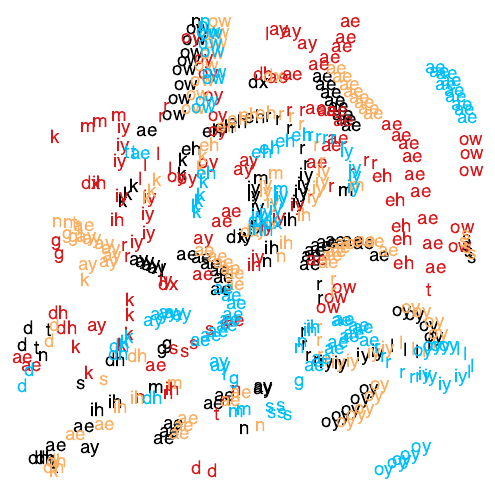}
        \vspace{-18pt}
        \caption{Layer 3}
    \end{subfigure}
    \begin{subfigure}[b]{0.300\linewidth}
        \centering
        \includegraphics[width=\linewidth]{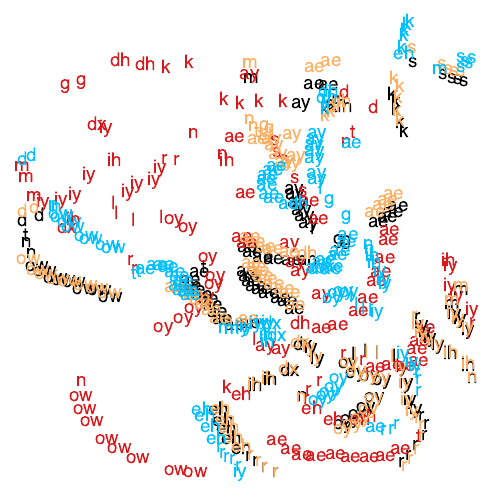}
        \vspace{-18pt}
        \caption{Layer 4}
    \end{subfigure}
    \hfill
    \begin{subfigure}[b]{0.300\linewidth}
        \centering
        \includegraphics[width=\linewidth]{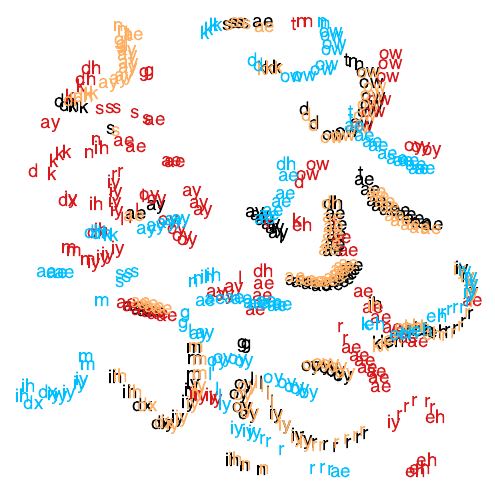}
        \vspace{-18pt}
        \caption{Layer 5}
    \end{subfigure}
    \hfill
    \begin{subfigure}[b]{0.300\linewidth}
        \centering
        \includegraphics[width=\linewidth]{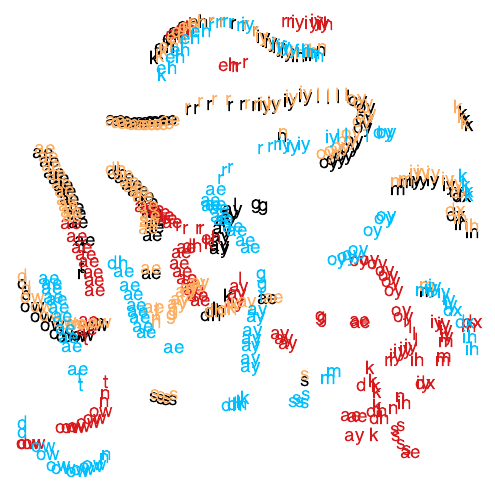}
        \vspace{-18pt}
        \caption{Layer 6}
    \end{subfigure}
    \begin{subfigure}[b]{0.300\linewidth}
        \centering
        \includegraphics[width=\linewidth]{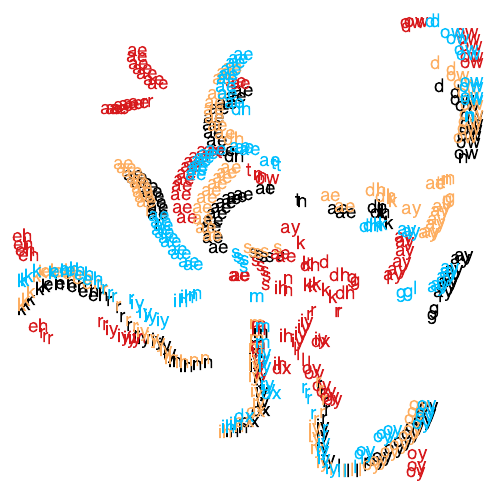}
        \vspace{-18pt}
        \caption{Layer 7}
    \end{subfigure}
    \hfill
    \begin{subfigure}[b]{0.300\linewidth}
        \centering
        \includegraphics[width=\linewidth]{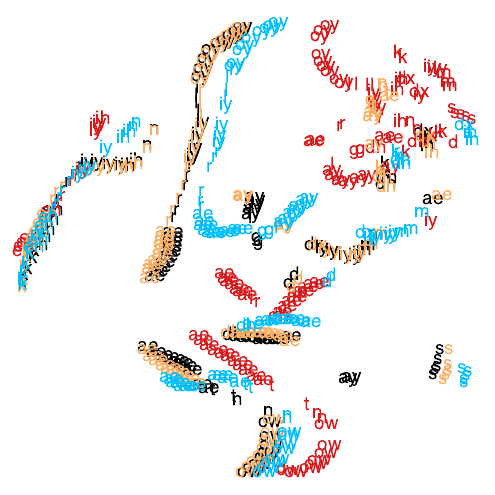}
        \vspace{-18pt}
        \caption{Layer 8}
    \end{subfigure}
    \hfill
    \begin{subfigure}[b]{0.300\linewidth}
        \centering
        \includegraphics[width=\linewidth]{fig/hubert_traj_cross_domain/hubert_L9.pdf}
        \vspace{-18pt}
        \caption{Layer 9}
    \end{subfigure}
    \begin{subfigure}[b]{0.300\linewidth}
        \centering
        \includegraphics[width=\linewidth]{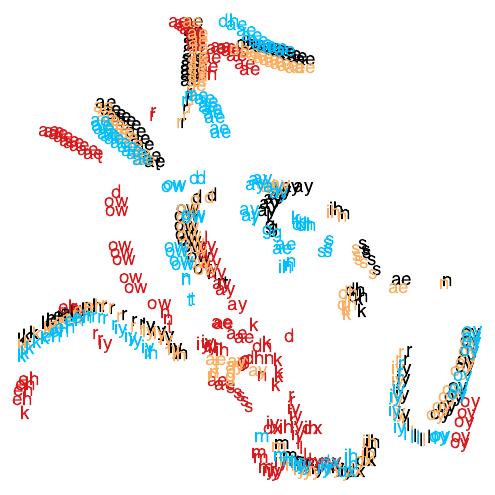}
        \vspace{-18pt}
        \caption{Layer 10}
    \end{subfigure}
    \hfill
    \begin{subfigure}[b]{0.300\linewidth}
        \centering
        \includegraphics[width=\linewidth]{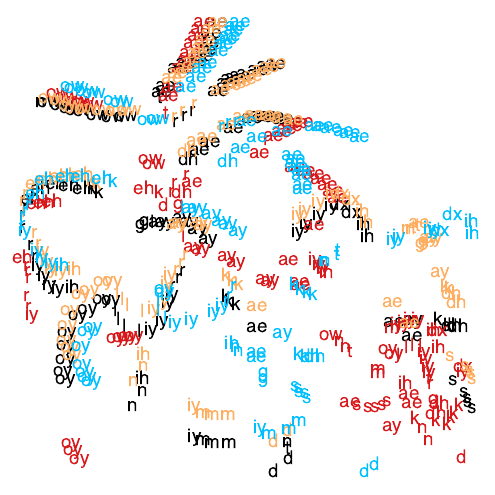}
        \vspace{-18pt}
        \caption{Layer 11}
    \end{subfigure}
    \hfill
    \begin{subfigure}[b]{0.300\linewidth}
        \centering
        \includegraphics[width=\linewidth]{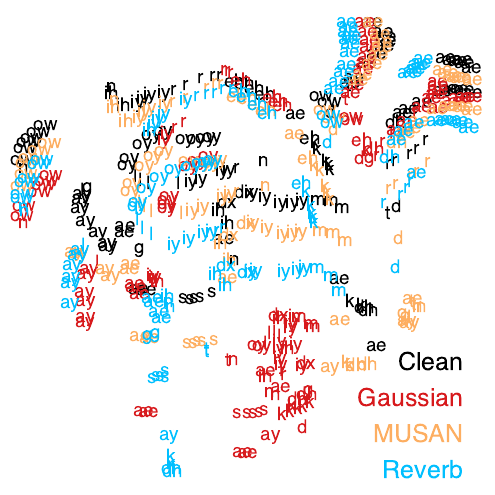}
        \vspace{-18pt}
        \caption{Layer 12}
    \end{subfigure}
    \caption{
        t-SNE visualization of HuBERT representations of the same utterance under different distortions~(see Fig.~\ref{fig:trajectory-domain} for details).
    }
    \label{fig:trajectory-hubert-domain}
    \vspace{-6pt}
\end{figure*}

%% file: fig/trajectory_rspin_domain_full.tex
\begin{figure*}
    \centering
    \begin{subfigure}[b]{0.300\linewidth}
        \centering
        \includegraphics[width=\linewidth]{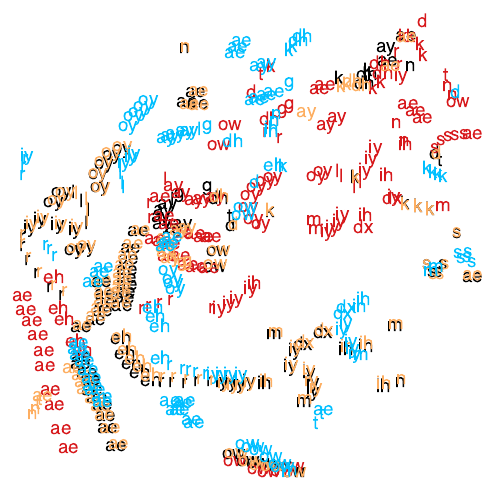}
        \vspace{-18pt}
        \caption{Layer 1}
    \end{subfigure}
    \hfill
    \begin{subfigure}[b]{0.300\linewidth}
        \centering
        \includegraphics[width=\linewidth]{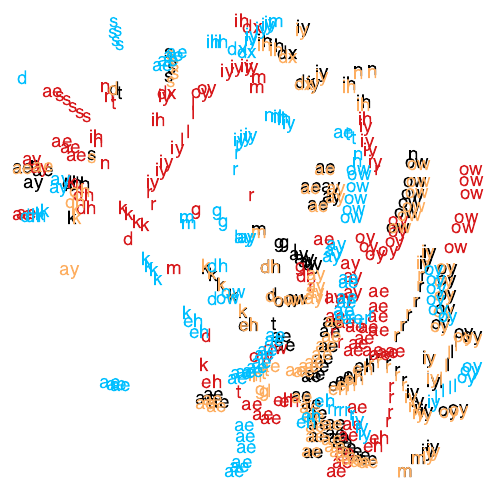}
        \vspace{-18pt}
        \caption{Layer 2}
    \end{subfigure}
    \hfill
    \begin{subfigure}[b]{0.300\linewidth}
        \centering
        \includegraphics[width=\linewidth]{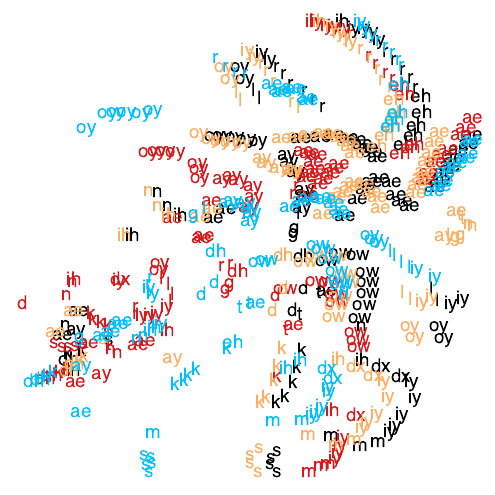}
        \vspace{-18pt}
        \caption{Layer 3}
    \end{subfigure}
    \begin{subfigure}[b]{0.300\linewidth}
        \centering
        \includegraphics[width=\linewidth]{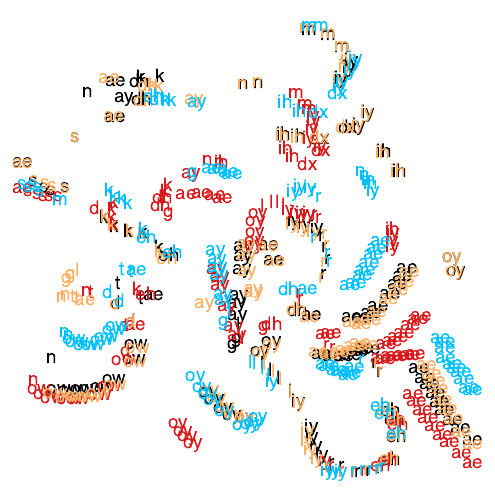}
        \vspace{-18pt}
        \caption{Layer 4}
    \end{subfigure}
    \hfill
    \begin{subfigure}[b]{0.300\linewidth}
        \centering
        \includegraphics[width=\linewidth]{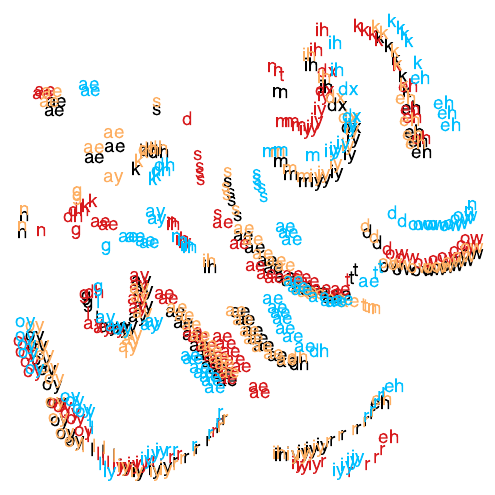}
        \vspace{-18pt}
        \caption{Layer 5}
    \end{subfigure}
    \hfill
    \begin{subfigure}[b]{0.300\linewidth}
        \centering
        \includegraphics[width=\linewidth]{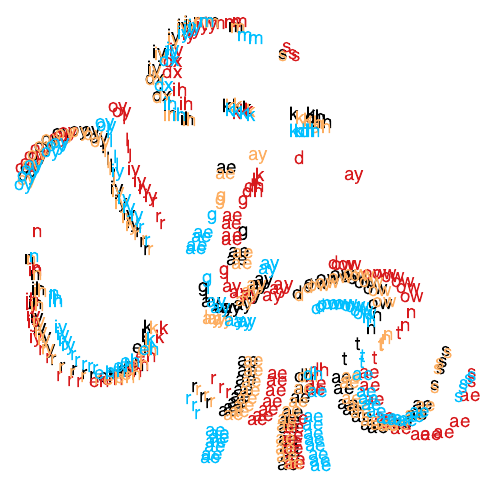}
        \vspace{-18pt}
        \caption{Layer 6}
    \end{subfigure}
    \begin{subfigure}[b]{0.300\linewidth}
        \centering
        \includegraphics[width=\linewidth]{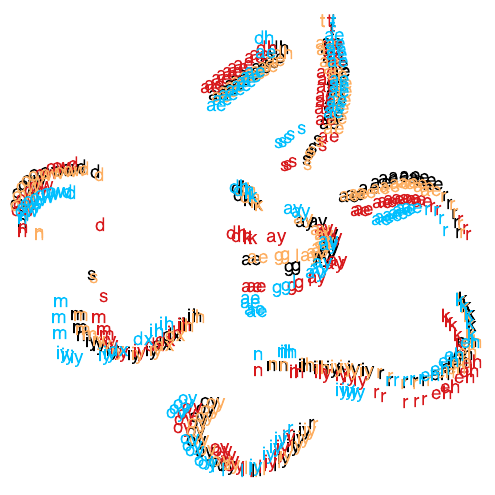}
        \vspace{-18pt}
        \caption{Layer 7}
    \end{subfigure}
    \hfill
    \begin{subfigure}[b]{0.300\linewidth}
        \centering
        \includegraphics[width=\linewidth]{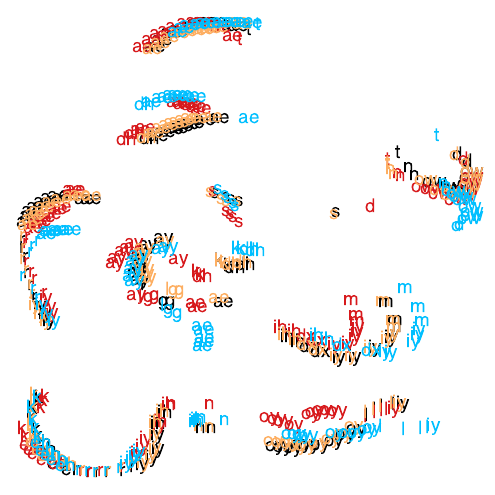}
        \vspace{-18pt}
        \caption{Layer 8}
    \end{subfigure}
    \hfill
    \begin{subfigure}[b]{0.300\linewidth}
        \centering
        \includegraphics[width=\linewidth]{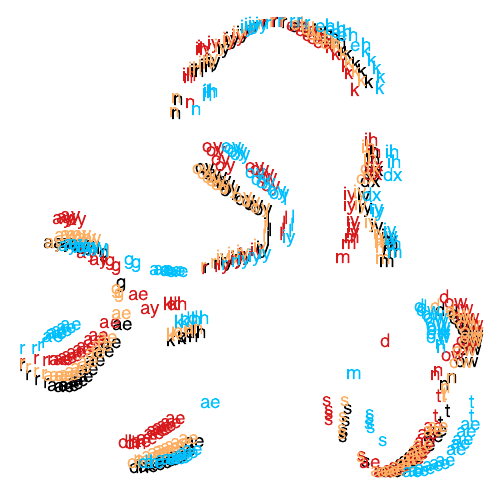}
        \vspace{-18pt}
        \caption{Layer 9}
    \end{subfigure}
    \begin{subfigure}[b]{0.300\linewidth}
        \centering
        \includegraphics[width=\linewidth]{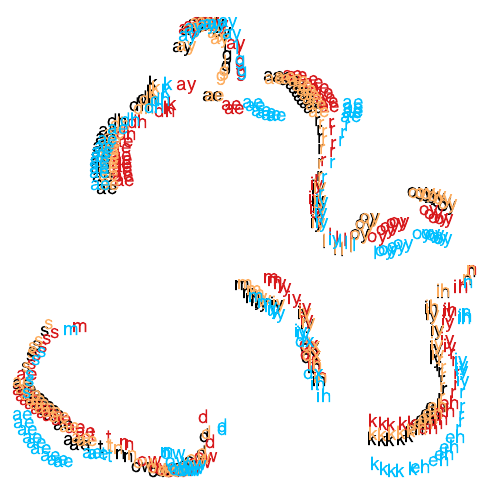}
        \vspace{-18pt}
        \caption{Layer 10}
    \end{subfigure}
    \hfill
    \begin{subfigure}[b]{0.300\linewidth}
        \centering
        \includegraphics[width=\linewidth]{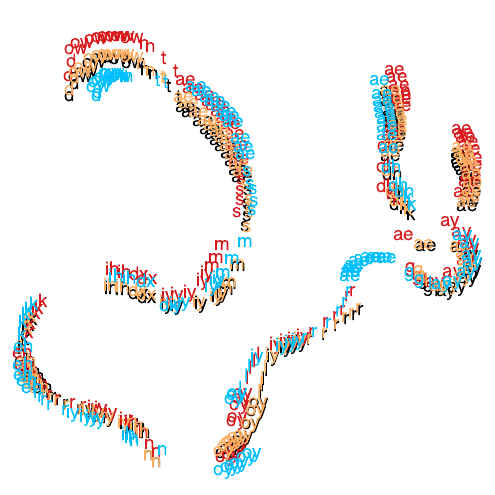}
        \vspace{-18pt}
        \caption{Layer 11}
    \end{subfigure}
    \hfill
    \begin{subfigure}[b]{0.300\linewidth}
        \centering
        \includegraphics[width=\linewidth]{fig/rspin_traj_cross_domain/spin2_hubert_v48_L12.pdf}
        \vspace{-18pt}
        \caption{Layer 12}
    \end{subfigure}
    \caption{
        t-SNE visualization of HuBERT + R-Spin representations of the same utterance under different distortions~(see Fig.~\ref{fig:trajectory-domain} for details).
    }
    \label{fig:trajectory-rspin-domain}
    \vspace{-6pt}
\end{figure*}